\documentclass{article}
\usepackage[utf8]{inputenc}
\usepackage[backref=page]{hyperref}
\usepackage{natbib}
\usepackage{geometry}
\usepackage{graphicx}
\usepackage{booktabs}
\usepackage{amssymb}
\usepackage{microtype}
\usepackage{adjustbox}
\usepackage{subcaption}
\usepackage{float}

\usepackage{amsmath}
\usepackage{bm}
\usepackage[utf8]{inputenc} 
\usepackage[T1]{fontenc}    
\usepackage{hyperref}       
\usepackage{url}            
\usepackage{nicefrac}       
\usepackage{microtype}      
\usepackage{xcolor}         

\def\1{\bm{1}}
\def\eps{{\epsilon}}

\def\vtheta{{\bm{\theta}}}

\def\vm{{\bm{m}}}

\def\vp{{\bm{p}}}

\def\vv{{\bm{v}}}

\def\vx{{\bm{x}}}

\def\vz{{\bm{z}}}

\def\mI{{\bm{I}}}

\def\mM{{\bm{M}}}

\def\mU{{\bm{U}}}
\def\mV{{\bm{V}}}

\def\mX{{\bm{X}}}
\def\mY{{\bm{Y}}}
\def\mZ{{\bm{Z}}}

\DeclareMathAlphabet{\mathsfit}{\encodingdefault}{\sfdefault}{m}{sl}
\SetMathAlphabet{\mathsfit}{bold}{\encodingdefault}{\sfdefault}{bx}{n}

\def\sP{{\mathbb{P}}}
\def\sQ{{\mathbb{Q}}}

\newcommand{\softmax}{\mathrm{softmax}}

\newcommand{\norm}{\mathrm{renorm}}
\newcommand{\CE}{\mathrm{CrossEnt}}
\newcommand{\cent}{\mathrm{center}}
\newcommand{\sg}{\mathrm{sg}}

\DeclareMathOperator*{\argmin}{arg\,min}

\DeclareMathOperator{\CoSim}{CoSim}

\DeclareMathOperator{\NN}{NN}
\DeclareMathOperator{\relu}{ReLU}

\usepackage{tikz}

\usepackage{hyperref}
\usepackage{url}
\usepackage{multirow}
\usepackage{enumitem}

\usepackage{amsthm}
\usepackage{amssymb}
\usepackage{mathtools}
\theoremstyle{plain}

\usepackage{varwidth}
\usepackage{enumitem}
\usepackage{tikz}
\usetikzlibrary{positioning}
\usetikzlibrary{shapes}
\usetikzlibrary{fit}
\usetikzlibrary{calc}

\usepackage{listings}
\usepackage{xcolor}
\usepackage{courier}

\definecolor{codegreen}{rgb}{0,0.6,0}
\definecolor{codegray}{rgb}{0.5,0.5,0.5}
\definecolor{codeblack}{rgb}{0.,0.,0.}
\definecolor{codepurple}{rgb}{0.58,0,0.82}
\definecolor{backcolour}{rgb}{0.95,0.95,0.92}

\lstdefinestyle{mystyle}{
    backgroundcolor=\color{backcolour},   
    commentstyle=\color{codegreen},
    keywordstyle=\color{codeblack},
    numberstyle=\tiny\color{codegray},
    stringstyle=\color{codepurple},
    basicstyle=\ttfamily\footnotesize,
    breakatwhitespace=false,         
    breaklines=true,                 
    captionpos=b,                    
    keepspaces=true,                 
    numbers=left,                    
    numbersep=5pt,                  
    showspaces=false,                
    showstringspaces=false,
    showtabs=false,                  
    tabsize=2,
    aboveskip=0pt,
    belowskip=-3pt
}

\newcommand{\SummaryCard}[3]{
\draw[rounded corners=\cardroundingradius] (0,0) rectangle  (\textwidth,#1);
\fill[gray!40,rounded corners=\striproundingradius] (\strippadding,#1-\strippadding) rectangle (\textwidth-\strippadding*1cm,#1-\stripheight-\strippadding) node[rotate=90,black] {};
\node at (\textwidth*0.5,#1-\stripheight*0.5-\strippadding)  {#2};
\node[below] at (\textwidth*0.5,#1-\stripheight-\strippadding-\textpadding) {{\topsize #3}};
}

\newcommand{\Quote}[2]{
}

\usepackage{pifont}
\usepackage{tgbonum}
\usepackage{cleveref}
\usepackage[framemethod=TikZ]{mdframed}
\usepackage{tikz}
\usetikzlibrary{positioning,calc}
\usetikzlibrary{shapes,arrows,chains}
\usepackage[linesnumbered,ruled,vlined]{algorithm2e}

\usepackage{enumitem}
\setlist[itemize]{align=parleft,left=0pt..1em}

\usepackage{hyperref}
\hypersetup{%
    colorlinks,
    linkcolor=purple,
    citecolor=gray,
    urlcolor=purple
}

\title{A Cookbook of Self-Supervised Learning}

\usepackage{authblk}

\author[*]{Randall Balestriero}
\author[*]{Mark Ibrahim}
\author[*]{\em Vlad Sobal}
\author[*]{\em Ari Morcos}
\author[*]{\em Shashank Shekhar}
\author[$\dagger$]{\em Tom Goldstein}
\author[*$\ddagger$]{\em Florian Bordes}
\author[*]{\em Adrien Bardes}
\author[*]{\em Gregoire Mialon}
\author[*]{\em Yuandong Tian}
\author[$\dagger$]{\em Avi Schwarzschild}
\author[**]{\em Andrew Gordon Wilson}
\author[$\dagger$]{\em Jonas Geiping}
\author[*$\S$]{\em Quentin Garrido}
\author[*$^\star$]{\em Pierre Fernandez}
\author[*]{\em Amir Bar}
\author[+]{\em Hamed Pirsiavash}
\author[*]{\em Yann LeCun}
\author[**]{\em Micah Goldblum}

\affil[*]{Meta AI, FAIR}
\affil[**]{New York University}
\affil[$\dagger$]{University of Maryland}
\affil[+]{University of California, Davis}
\affil[$\ddagger$]{Universite de Montreal, Mila}
\affil[$\S$]{Univ Gustave Eiffel, CNRS, LIGM}
\affil[$\star$]{Univ. Rennes, Inria, CNRS, IRISA}
\affil[italic]{Equal contributions, randomized ordering}
\date{}

\begin{document}

\pgfmathsetmacro{\cardroundingradius}{3mm}
\pgfmathsetmacro{\striproundingradius}{2mm}
\pgfmathsetmacro{\cardwidth}{5}
\pgfmathsetmacro{\cardheight}{8}
\pgfmathsetmacro{\stripheight}{0.4}
\pgfmathsetmacro{\strippadding}{0.1}
\pgfmathsetmacro{\textpadding}{0.2}
\pgfmathsetmacro{\ruleheight}{0.05}
\newcommand{\topsize}{\footnotesize}
\newcommand{\bottomsize}{\tiny}

\maketitle

\newpage

\tableofcontents

\newpage

\section{What is Self-Supervised Learning and Why Bother?}

\textit{Self-supervised learning}, dubbed “the dark matter of intelligence” \footnote{\url{https://ai.facebook.com/blog/self-supervised-learning-the-dark-matter-of-intelligence/}}, is a promising path to advance machine learning.
As opposed to \textit{supervised learning}, which is limited by the availability of labeled data, self-supervised approaches can learn from vast unlabeled data \citep{chen2020simple,misra2020self}. 
Self-supervised learning (SSL) underpins deep learning's success in natural language processing leading to advances from automated machine translation to large language models trained on web-scale corpora of unlabeled text \citep{brown2020language,popel2020transforming}.
In computer vision, SSL pushed new bounds on data size with models such as SEER trained on 1 billion images \citep{goyal2021self}. SSL methods for computer vision have been able to match or in some cases surpass models trained on labeled data, even on highly competitive benchmarks like ImageNet \citep{tomasev2022pushing, he2020momentum, deng2009imagenet}. SSL has also been successfully applied across other modalities such as video, audio, and time series \citep{wickstrom2022mixing,liu2022audio,schiappa2022self}.

Self-supervised learning defines a pretext task based on unlabeled inputs to produce descriptive and intelligible representations \citep{hastie2009overview,goodfellow2016deep}.
In natural language, a common SSL objective is to mask a word in the text and predict the surrounding words. This objective of predicting the context surrounding a word encourages the model to capture relationships among words in the text without the need for any labels. 
The same SSL model representations can be used across a range of downstream tasks such as translating text across languages, summarizing, or even generating text, along with many others.
In computer vision, analogous objectives exist with models such as MAE or BYOL learning to predict masked patches of an image or representation \citep{grill2020bootstrap,he2022masked}. Other SSL objectives encourage two views of the same image, formed by say adding color or cropping, to be mapped to similar representations.

With the power to train on vast unlabeled data comes many benefits. While traditional supervised learning methods are trained on a specific task often known a priori based on the available labeled data, SSL learns generic representations useful across many tasks. 
SSL can be especially useful in domains such as medicine where labels are costly or the specific task can not be known a priori \citep{krishnan2022self,CIGA2022100198}.
There's also evidence SSL models can learn representations that are more robust to adversarial examples, label corruption, and input perturbations---and are more fair---compared to their supervised counterparts \citep{hendrycks2019using, goyal2022vision}. 
Consequently, SSL is a field garnering growing interest. 
Yet, much like cooking, training SSL methods is a delicate art with a high barrier to entry.

\subsection{Why a Cookbook for Self-Supervised Learning?}

While many components of SSL are familiar to researchers, successfully training a SSL method involves a dizzying set of choices from the pretext tasks 
to training hyper-parameters. 
SSL research has a high barrier to entry due to (i) its computational cost, (ii) the absence of fully transparent papers detailing the intricate implementations required to fully enable SSL's potential, and (iii) the absence of a unified vocabulary and theoretical view of SSL.
As SSL established a distinct paradigm from traditional {\em reconstruction-based} unsupervised learning methods such as (denoising, variational) Autoencoders \citep{vincent2008extracting,vincent2010stacked,kingma2013auto}, our vocabulary for understanding SSL in a unified view is limited.
In fact, attempts at unifying SSL methods under a single viewpoint have only started to emerge in the last year \citep{haochen2021provable,balestriero2022contrastive,shwartz2022we,garrido2022duality}.
Without a common ground to characterize the different components of SSL methods, it's more challenging for researchers to start working on SSL methods. 
Meanwhile, SSL research is in dire need for new researchers since SSL is now deployed throughout the real-world. 
Yet, many open research questions remain regarding SSL's generalization guarantees, fairness properties, and robustness to adversarial attacks or even naturally occurring variations. 
Such questions are crucial to the reliability of SSL methods.

Furthermore, SSL---which is empirically driven---comes with many moving pieces (mostly hyper-parameters) that may impact key properties of the final representations and are not necessarily well-detailed in published work. That is, to start studying SSL methods, one must first exhaustively empirically probe those methods to fully grasp the impact and behaviors of all those moving pieces. 
Such empirical blind spots are strong limitations as they demand large computational resources and pre-existing hands-on experience. 
All in all, the co-occurrence of SOTA performances from seemingly different yet overlapping methods, little existing theoretical research, and widespread real-world deployment, make the need for 
a cookbook unifying the techniques and their recipes essential to lower SSL's research barrier to entry.

Our goal is to lower the barrier to entry into SSL research by laying the foundations and latest SSL recipes in the style of a cookbook.
To successfully cook, you must first learn the basic techniques: chopping, saut\'{e}ing, etc.
We begin in \Cref{sec:methods} with the fundamental techniques of self-supervised learning using a common vocabulary.
Specifically, we describe the families of methods along with theoretical threads to connect their objectives in a unified perspective.
We highlight key concepts such as loss terms or training objectives in concept boxes.
Next, a cook must learn to skillfully apply the techniques to form a delicious dish. 
This requires learning existing recipes, assembling ingredients, and evaluating the dish.
In \Cref{sec:practical_matters} we introduce the practical considerations to implementing SSL methods successfully. 
We discuss common training recipes including hyperparameter choices, how to assemble components such as architectures and optimizers, as well as how to evaluate SSL methods. 
We also share practical tips from leading researchers on common training configurations and pitfalls.
We hope this cookbook serves as a practical foundation for successfully training and exploring 
self-supervised learning.

\section{The Families and Origins of SSL}
\label{sec:methods}

SSL methods have enjoyed a renaissance since 2020, thanks in large part to the availability of extremely large datasets and high-memory GPUs. However, the origins of SSL go back to the very beginning of the deep learning era. 

\subsection{Origins of SSL}
Contemporary methods build upon the knowledge we gained from early experiments. In this section, we give a brief overview of the main ideas of SSL prior to 2020. While many of the specific methods have fallen out of mainstream use because they no longer provide state-of-the-art performance on benchmark problems, and they will not be discussed in great detail, 
the ideas from these papers form the foundation for many of the modern methods. 
For example, the core objective of restoring missing or distorted parts of an input or contrasting two views of the same image form the foundation for modern SSL methods.
Early progress in SSL focused on the development of methods that fell into the following (sometimes overlapping) categories:

\vspace{4pt}
{\bf 1. Information restoration:}
A wide range of methods have been developed that mask or remove something from an image, and then train a neural network to restore the missing information.   
Colorization-based SSL methods convert an image to grayscale, and then train a network to predict the original RGB values \citep{zhang2016colorful,larsson2016learning,vondrick2018tracking}. Because colorization requires understanding object semantics and boundaries, colorization was demonstrated as an early SSL method for object segmentation.
The most straightforward application of information restoration is to mask, aka remove, a portion of an image and then train a network to inpaint the missing pixel values \citep{pathak2016context}. This idea evolved into masked auto-encoding methods \citep{he2022masked}, in which the masked region is a union of image patches that can be predicted using a transformer.

\vspace{4pt}
{\bf 2. Using temporal relationships in video:}
While the focus of this review is on image (and not video) processing, a range of specialized methods have been developed for learning single-image representations by pre-training on videos. Note that information restoration methods are particularly useful for videos, which contain multiple modalities of information that can be masked.
\citet{wang2015unsupervised} pre-train a model using a triplet loss that promotes similarities between representations of an object in two different frames. The resulting model performed well for object detection. \citet{pathak2017learning} trains a model to predict the motion of objects in a single frame, and adapts the resulting features to solve single-frame detection problems. \citet{agrawal2015learning} predicts the ego-motion of a camera given multiple frames. \citet{owens2016ambient} propose to remove the audio track from a video, and then predict the missing sound.  For specialized applications like depth mapping, self-supervised methods have been proposed that learn monocular depth models from unlabeled image pairs \citep{eigen2014depth} and later the frames from a single-camera video \citep{zhou2017unsupervised}. Such methods remain an active area of research. 

\vspace{4pt}
{\bf 3. Learning spatial context:}
This category of methods trains a model to understand the relative positions and orientations of objects within a scene.  RotNet \citep{gidaris2018unsupervised} masks the direction of gravity by applying a random rotation and then asks the model to predict the rotation. \citet{doersch2015unsupervised} is one of the first SSL methods that simply predicts the relative location of two randomly sampled patches in an image. This strategy was superseded by ``jigsaw'' methods \citep{pathak2016context,noroozi2018boosting} that break an image into an array of disjoint patches and predict the relative location of each. A different spatial task is learning to count \citep{noroozi2017representation}: the model is trained to output the number of objects in an image in a self-supervised way.

\vspace{4pt}
{\bf 4. Grouping similar images together:}
One can learn rich features by grouping semantically similar images together. K-means clustering is one of the most widely used methods from classical machine learning.  A number of studies have adapted k-means to perform SSL with neural models.
Deep clustering alternates between assigning labels to images by performing k-means in the feature space, and updating the model to respect these assigned class labels \citep{caron2018deep}. More recent treatments of this approach use mean-shift updates to push features towards their cluster center, and have been shown to complement BYOL, a method based on two networks with the objective to predict pseudo-labels for each sample \citep{koohpayegani2021mean} (discussed in Section \ref{sec:self-distillation}). Other improvements to deep clustering include using optimal transport methods in feature space to create more informative clusters \citep{asano2019self}.

\vspace{4pt}
{\bf 5. Generative models:}
An early influential SSL method is greedy layer-wise pretraining \citep{bengio2006greedy}, in which layers of a deep network are trained one-at-a-time using an autoencoder loss.  An analogous approach from the time used Restricted Boltzman Machines (RBMs), which could be trained layer-wise and stacked to create deep belief nets \citep{hinton2006fast}.
While these methods were abandoned in favor of simpler initialization strategies and longer training runs, they were historically impactful uses of SSL, as they enabled the training of the first ``deep'' networks. Later advancements improved on the representation learning ability of auto-encoders, including denoising autoencoders \citep{vincent2008extracting}, cross-channel prediction
\citep{zhang2017split}, and deep canonically correlated autoencoders \citep{wang2015deep}.
Nonetheless, it was ultimately found that representation transferability is better when the auto-encoder is asked to restore a missing part of its input, resulting in the ``information restoration'' category of SSL methods.

Generative Adversarial Networks (GANs) \citep{goodfellow2014generative} consist of an image generator and a discriminator that differentiates real images from generated images. Both components of this model pair can be trained without supervision, and both potentially contain knowledge useful for transfer learning. Early GANs papers \citep{salimans2016improved} experimented with downstream image classification using GAN components.  Specialized feature learning routines have also been developed that modify the discriminator \citep{springenberg2015unsupervised}, add a generator \citep{dai2017good}, or learn additional mappings from image to latent space \citep{donahueadversarial} to improve transfer learning.

\vspace{4pt}
{\bf 6. Multi-view invariance:}
Many modern SSL methods, especially those that we focus on in this article, use contrastive learning to create feature representations that are invariant to simple transforms. The idea of contrastive learning is to encourage a model to represent two augmented versions of an input similarly. A number of methods led the charge in this direction by enforcing invariance in various ways before contrastive learning was widely adopted. 

One of the most popular frameworks for learning from unlabeled data is to use a weakly trained network to apply pseudolabels to images, and then train using these labels in a standard supervised fashion \citep{lee2013pseudo}. This approach was later improved by enforcing invariance to transformations.
Virtual adversarial training \citep{miyato2018virtual} trains a network on images using their pseudolabels, and additionally performs adversarial training so that learned features are nearly invariant to small perturbations to the input image.  Later works focused on maintaining invariance to data augmentation transforms.  Important early methods in this category include MixMatch \citep{berthelot2019mixmatch}, which chooses pseudolabels by averaging outputs of a network on several different random augmentations of the training images, resulting in labels that are augmentation invariant.
Around the same time, it was discovered that good SSL performance could be achieved by training a network to maximize the mutual information between the representations of an image under different views \citep{bachman2019learning}.  These augmentation-based methods formed a bridge between the older methods described above and the contemporary methods that are the focus of this paper.

With these origins, we now turn to categorizing SSL into four broad families: The Deep Metric Learning Family, The Self-Distillation Family, The Canonical Correlation Analysis Family, and the Masked Image Modeling Family.

\subsection[The Deep Metric Learning Family: SimCLR/NNCLR/MeanSHIFT/SCL]{The Deep Metric Learning Family:\\ SimCLR/NNCLR/MeanSHIFT/SCL}

The Deep Metric Learning (DML) family of methods is based on the principle of encouraging similarity between semantically transformed versions of an input.
DML originated with the idea of a \textit{contrastive loss}, which transforms this principle into a learning objective.
Contrastive loss was first introduced in  \citep{bromley1993signature} then more formally defined in \citep{chopra2005learning,hadsell2006dimensionality}. 
In DML one trains a network to predict whether two inputs are from the same class (or not) by making their embedding close (or far from each other).
Since data is without labels, to identify similar inputs, we often form variants of a single input using known semantic preserving transformations. 
The variants of the inputs are called \textit{positive pairs} or examples; the samples we wish to make dissimilar are called \textit{negatives}. 
Often there's a margin parameter, $m$, imposing the distance between examples from different classes should be larger than $m$. 
Similar to the contrastive loss, the Triplet loss \citep{weinberger2009distance,chechik2010large,schroff2015facenet} shares a similar spirit, but is composed of triplets: a query, a positive example, and a negative example (see \cref{eq:triplet}).
Compared to contrastive loss, triplet loss only requires the difference of (dis-)similarities between positive and negative examples to the query point to be larger than a margin $m$. 

The shift from DML to what is now referred to as SSL might have occurred when 
\citet{sohn2016improved} introduced the (N+1)-tuple loss, a loss similar to the contrastive predictive
coding (CPC) loss from \citep{oord2018representation}. 
The use of other sample positive views as the negative view of other pairs is introduce as an efficient strategy coined {\em N-pair-mc loss}. 
\citet{ni2021close} shows that contrastive learning is a special case of meta-learning, and existing meta-learners can be directly applied to SSL with competitive performance.
CPC was extended to images in \citep{henaff2020data}.
A key ingredient in CPC was the introduction of the InfoNCE loss described in \ref{fig:infonce_plus} \cite{oh2016deep}, which became central in SSL. 

To summarize, the main paradigm shift between DML and Contrastive SSL arises from a few key changes, 
namely using data-augmentation 
instead of sampling to obtain the positive/negative pairs, the use of deeper networks, and the use of a predictor network, which we 
note in \Cref{fig:dml_ssl}. One of the most prominent methods coming from the paradigm shift to SSL in the deep learning family is SimCLR.

\textbf{SimCLR} learns visual representations by encouraging similarity between two augmented views of an image. In SimCLR the two views are formed by applying a combination of transformations including random resizing, cropping, color jittering, and random blurring. After encoding each view, SimCLR uses a \textit{projector}, often a MLP (multi-layer perceptron) followed by a ReLU (rectified linear unit) activation, to map the initial embeddings into another space where the constrative loss if applied to encourage similarity between the views. For downstream tasks, extracting the representation before the projector has been shown to improve performance. Further discussions of the role of the projector are in sections \ref{sec:projector_theory} and \ref{sec:projector}. 

Another key ingredient along with the InfoNCE loss used in SimCLR is the non-parametric softmax introduced by \citet{wu2018unsupervised}. This name is motivated by removing the need to have a "parametrized" linear layer on top of the representation to compute the softmax by instead comparing representations with each others. 
This loss formulation already contained a \textit{temperature parameter} in the softmax which is responsible for increasing or decreasing the sharpness of events in predictions.
Other noteworthy developements include \citet{schroff2015facenet} use triplet loss with active triplet selection (hard positive, hard negative) either online from the current mini-batch or from a past checkpoint akin to momentum networks (discussed in section \ref{sec:self-distillation}).
\citet{weinberger2009distance} introduced push-pull weighting, to push negatives apart while pulling positives together, in a triplet loss to increase the margin of K-NN based models.
\citet{tian2020contrastive} introduced the possibility of many positive views.

Aside from forming positives using semantic preserving transformations, mining positive pairs naturally arising in data is also possible. An iconic
triplet loss is based on video frames where the positive pairs come from nearby frames (while negatives are from far away frames) developed in \citet{sermanet2018time} coined Time-Contrastive (TC)
Time-Contrastive Learning. Nonlinear ICA \citep{hyvarinen2016unsupervised} introduced a proof that you can learn the log PDF when doing classification tasks.
\citet{alexey2015discriminative} trains a classification pretext task by transforming image patches in comparison to different transformations of image patches. 
One disadvantage is that this setup can involve too many classes leading performance to degrade on downstream tasks.
To overcome this,  NCE has been successfully employed in \citet{mnih2012fast,mnih2013learning} to modify the denominator in order not to loop over all classes. 
This is an alternative to sampling based estimation of the gradient that was found to be less stable \citep{bengio2003quick,bengio2008adaptive}. 
This introduces the concept of will become momentum encoder by imposing that features maps do not vary quickly referred to as proximal algorithm \citep{parikh2014proximal}. 
One other consideration in SSL motivated by the DML is the idea of ``hard negative data mining'' where the negative samples are intentionaly selected to be close to but distinct from the positives to form a more challenging learning objective. Next we describe an alternative to deep metric learning based on self-distillation.

\begin{figure}[t!]
\centering
\begin{tikzpicture}
    \SummaryCard{9.5}{Noise Contrastive Estimation: Learning Unnormalized Densities}{\begin{minipage}{0.95\textwidth}
    \begin{itemize}
        \item introduced by \citet{gutmann2010noise} to learn unnormalized probability distributions given i.i.d observations $\vx_1,\dots,\vx_N$ from the distribution $X\sim p_{X}$. NCE enables approximation of $p_{X}$ by a parametrized function $f_{\theta}$ without enforcing $\int f_{\theta}(\vx)d\vx=1$ during training
        \item let's first introduce a noise variable $\epsilon \sim p_{\epsilon}$ and let's consider the following mixture distribution
        $$T\sim \mathcal{B}(s),s \in (0,1)$$ $$ V \sim X1_{\{T=1\}}+\varepsilon 1_{\{T=0\}}$$
        \item using Bayes rule and denoting $\eta=(1-s)/s$ we have $$p_{T|V}(T=1|V=\vv)=\frac{p_{V|T}(V=\vv|T=1)}{p_{V|T}(V=\vv|T=1)+\eta p_{V|T}(V=\vv|T=0)},$$
        \item parametrize $p_{V|T}(V=\vv|T=1)=f_{\theta}(\vv)\exp(c)$ with $f_{\theta}>0$ and learnable parameters $\{\theta,c\}$
        \item minimize the NLL of logistic regression (usual binary classification set-up)
        \begin{align*}
            \mathcal{L}(\theta,c)=-\mathbb{E}_{(\vv,t)\sim(V,T)}\log[p_{T|V}(T=t|V=\vv)]
        \end{align*}
        \item the minimum is attained at $f_{\theta^*}\exp(c^*)=p_{X}$ if $p_{X}(\vv)=0\implies p_{\epsilon}(\vv)>0$. If $f_{\theta}$ is powerful enough, one can set $c=0$ and the model will self-normalize \citep{mnih2012fast}
        \item \citet{ceylan2018conditional} extends NCE to nonindependent noise realization i.e. $\epsilon$ depends on $X$, \citet{ma2018noise} considers conditional distribution $X|Y$, \citet{dyer2014notes} compares NCE and Negative Sampling \citep{mikolov2013distributed} (the latter being a special case of the former) both extending Importance Sampling estimation \citep{bengio2003quick} of the partition function (normalization factor)
        \end{itemize}
    \end{minipage}
    }
\end{tikzpicture}
\caption{Noise Contrastive Estimation}
\label{fig:history_NCE}
\end{figure}

\begin{figure}[t!]
\centering
\begin{tikzpicture}
    \SummaryCard{14}{A Brief History of the infoNCE loss}{\begin{minipage}{0.95\textwidth}
    In the descriptions below $z_i$ denotes the model representation of sample i, $\sP$ denotes the set of positive samples, and $\tau$ is the temperature hyperparameter.
    
    \begin{itemize}
        \item \citet{bromley1993signature,chopra2005learning} introduces the {\bf contrastive loss} for Deep Metric Learning
        \begin{align}
            \mathcal{L}_{\rm cont}(\mZ)=\sum_{(i,j)\in \sP}\|\vz_j-\vz_i\|_2+\sum_{(i,j)\not \in \sP}\relu(m-\|\vz_i-\vz_j\|_2)^2,m>0,\label{eq:contrastive}
        \end{align}
        \item \citet{goldberger2004neighbourhood} introduced {\bf Neighbourhood Component Analysis} to improve maximum margin of NN-classifiers by learning a quadratic distance (Mahalanobis distance is a special case of such a distance) using
        \begin{align}          
            \mathcal{L}_{\rm NCA}(\mZ)=-\sum_{(i,j)\in\sP}\frac{e^{-\|\vz_i-\vz_j\|_2^2}}{\sum_{(k,l)\in[N]^2}e^{-\|\vz_k-\vz_l\|_2^2}},
        \end{align}
        \item \citet{weinberger2009distance,chechik2010large} extends \cref{eq:contrastive} to a {\bf triplet loss}
        \begin{align}
            \mathcal{L}_{\rm triplet}(\mZ)=\sum_{(i,j) \in \sP}\sum_{(k,l)\not \in \sP,k=i\}}\relu(\|\vz_i-\vz_j \|_2-\|\vz_i-\vz_k \|+m),m>0,\label{eq:triplet}
        \end{align}
        \item \citet{sohn2016improved} extends the triplet and NCA losses to form the {\bf (N+1)-tuple loss}
        \begin{align}          
            \mathcal{L}_{\rm tuple}(\mZ)=-\sum_{(i,j)\in\sP}\log\left(\frac{e^{\langle\vz_i,\vz_j\rangle}}{\sum_{(k,l)\in \sP}e^{\langle\vz_i,\vz_l\rangle}}\right)+\beta \|\mZ\|_F^2,
        \end{align}
        where the denominator sum only runs through one view of the other samples, and the negative distance is replaced by the inner product \underline{and} $\ell_2$-penalty of the feature maps $\mZ$. Explicit normalization was found to be unstable yet introduced (along with a temperature parameter) in \citet{yu2019deep}.
        \item \citet{wu2018unsupervised} introduces the {\bf Noise-Contrastive Estimation} (NCE) loss without positive pairs
        \begin{align}          
            \mathcal{L}_{\rm NCE}=-\sum_{n=1}^{N}\log\left(\frac{e^{\CoSim( \vz_i,\vz_i^{(t-1)})/\tau}}{\sum_{k=1}^{N}e^{\CoSim(\vz_i,\vz_k)/\tau}}\right)+\beta \|\mZ-\mZ^{(t-1)}\|_F^2,\label{eq:wu_nce}
        \end{align}
        also coining the term non-parametric softmax. NCE loss introduces explicit normalization, a temperature parameter $\tau$, and the idea of momentum encoder (via proximal optimization method), and employs NCE to approximate the denominator when $N$ is large
        \item \citet[{\bf CPC}]{oord2018representation} coins the name {\bf infoNCE} by removing the proximal constraint and using positive pairs
        \begin{align}          
            \mathcal{L}_{\rm infoNCE}=-\sum_{(i,j)\in\sP}\log\left(\frac{e^{\CoSim( \vz_i,\vz_j)/\tau}}{\sum_{k=1}^{N}e^{\CoSim(\vz_i,\vz_k)/\tau}}\right),
        \end{align}
        \end{itemize}
    \end{minipage}
    }
\end{tikzpicture}
\caption{History of infoNCE}
\label{fig:history_infonce}
\end{figure}

\begin{figure}[t!]
\centering
\begin{tikzpicture}
    \SummaryCard{10}{The infoNCE Offsprings}{\begin{minipage}{0.95\textwidth}
    \begin{itemize}
        \item \citet[{\bf MoCo}]{he2020momentum} introduces momentum encoder as an alternative to the memory bank regularization of \cref{eq:wu_nce} and introduces a queue to store many negative samples from previous batches; \citep[{\bf MoCoV2}]{chen2020improved} adds a projector, \citep[{\bf MoCoV3}]{chen2021empirical} adds ViTs
        \item \citet[{\bf SimCLR}]{chen2020simple} removes the momentum encoder and the $i^{\rm th}$ term from the denominator coining it {\bf NT-Xent} (Normalized Temperature-scaled cross entropy)
        \begin{align*}          
            \mathcal{L}_{\rm NT-Xent}(\mZ)=-\sum_{(i,j)\in \sP}\frac{e^{\CoSim( \vz_i,\vz_j)}}{\sum_{k=1}^{N}\1_{\{k\not = i\}}e^{\CoSim(\vz_i,\vz_k)}},
        \end{align*}
        \item \citet[{\bf DCL}]{yeh2021decoupled} additionally removes the positive pair in the denominator
        \begin{align*}          
            \mathcal{L}_{\rm DCL}(\mZ)=-\sum_{(i,j)\in \sP}\frac{e^{\CoSim( \vz_i,\vz_j)}}{\sum_{k=1}^{N}\1_{\{k\not = i \wedge (i,k)\not = \sP\}}e^{\CoSim(\vz_i,\vz_k)}},
        \end{align*}
        \item \citet[{\bf NNCLR}]{dwibedi2021little} uses nearest neighbors from a queue $\sQ$
        \begin{align*}          
            \mathcal{L}_{\rm NNCLR}(\mZ)=-\sum_{(i,j)\in \sP}\frac{e^{\CoSim(\NN(\vz_i,\sQ), \vz_j)}}{\sum_{(k,l)\in \sP}^Ne^{\CoSim(\NN(\vz_i,\sQ),\vz_{l})}},
        \end{align*}
        \item \citet[{\bf RELIC}]{mitrovic2020representation} adds a regularization term to enforce invariance
        \begin{align*}          
            \mathcal{L}_{\rm RELIC}(\mZ)=-\sum_{(i,j)\in \sP}\frac{e^{\CoSim( \vz_i,\vz_j)}}{\sum_{k=1}^{N}\1_{\{k\not = i\}}e^{\CoSim(\vz_i,\vz_k)}}+KL(p(\vz_i),p(\vz_j)),
        \end{align*}
        \item \citet[{\bf PCL}]{li2020prototypical} uses prototypes 
        \end{itemize}
    \end{minipage}
    }
\end{tikzpicture}
\caption{Extensions of the infoNCE loss.}
\label{fig:infonce_plus}
\end{figure}

\begin{figure}[t!]
\centering
\begin{tikzpicture}
    \SummaryCard{4.8}{Paradigm Shift Between Deep Metric Learning and Contrastive SSL}{
    \def\arraystretch{1.5}
        \begin{tabular}{rcl}
             \multicolumn{1}{c}{\bf Deep Metric Learning} &  & \multicolumn{1}{c}{\bf Contrastive SSL}\\
             \parbox{6.2cm}{positive/negative pairs come from labels or fixed transforms e.g. two halves of an image}& $\implies$&\parbox{6.45cm}{positive pairs come from designed DAs that are continuously sampled, negative pairs are all non-positive pairs regardless of class membership}\\
             Hard-Negative Sampling for each mini-batch& $\implies$  & random sampling\\
             encoder DN & $\implies$ & encoder DN $+$ projector MLP\\ 
             small dataset (N<200k)& $\implies$&large dataset\\
             zero-shot k-NN validation & $\implies $&
             
             \parbox{5.5cm}{-zero-shot k-NN validation\\ -zero/few-shot/fine-tuning linear probing}
        \end{tabular}
    }
\end{tikzpicture}
\caption{Deep Metric Learning versus Contrastive SSL}
\label{fig:dml_ssl}
\end{figure}

\subsection[The Self-Distillation Family:
BYOL/SimSIAM/DINO]{The Self-Distillation Family:
BYOL/SimSIAM/DINO}
\label{sec:self-distillation}

\begin{figure}[t!]
\centering
\begin{tikzpicture}
    \SummaryCard{10.5}{A Brief History of the Self-Distillation Family}{\begin{minipage}{0.95\textwidth}
    \begin{itemize}
    \item \citet[{\bf MMC}]{xu2004maximum,joulin2010discriminative} searches pseudo-labels so that if a classifier were train on them it would have good margin (on true labels)
        \item \citet[{\bf NaT}]{bojanowski2017unsupervised} introduces Noise as Targets i.e. $C$ real {\em frozen} targets $\mM\triangleq [\vm_1,\dots,\vm_N] \in \mathbb{R}^{D \times C}$ with {\em assignment constraints} of $P\triangleq [\vp_1,\dots,\vp_N] \in \{0,1\}^{C \times N}$ with
        \begin{align}
            \mathcal{L}_{\rm NaT}=\min_{P:1P\leq 1,P^T1=1}-\sum_{n=1}^{N}CosSim(f_{\theta}(\vx_n),\mM \vp_n),
        \end{align}
        \item \citet[{\bf DeepCluster}]{caron2018deep} extends NaT by allowing learning of the targets in a K-means fashion with various cluster sampling and reallocation tricks to prevent collapse
        \begin{align}
            \mathcal{L}_{\rm DeepCluster}=CrossEntropy(f_{\theta}(\vx),\argmin_{k}\|f_{\theta}(\vx)-\vm_k\|_2^2) + K-means (f_{\theta}(\mX),\mM),
        \end{align}
        \item \citet[{\bf SLSC}]{YM.2020Self-labelling} further prevents collapse in DeepCluster through {\em constrained clustering membership} using Sinkhorn to infer the cluster membership probabilities
        \item \citet[{\bf BYOL}]{grill2020bootstrap} introduces BYOL removing the clustering step, introducing a {\em predictor} and projector network, defining the continuous targets as the output of a momentum network, renormalize each sample representation by its $\ell_2$-norm and leverage positive pairs. The predictor acts as a whitening operator preventing collapse \citep{tian2021understanding}, and momentum network can be applied only to the projector \citep{pham2022pros}
        \item \citet[{\bf SimSIAM}]{chen2021exploring}  replaces the BYOL moving average encoder by a stop-gradient
        \item \citet[{\bf DINO}]{caron2021emerging} introduces DINO which extends BYOL and SimSIAM to discrete representations/targets and still relies on momentum encoder
        \item \citet[{\bf iBOT}]{zhou2021ibot} and \citet[{\bf DINOv2}]{oquab2023dinov2} build upon DINO by combining its objective with a latent space masked-image modeling one, combining the best of both families 
        \end{itemize}
    \end{minipage}
    }
\end{tikzpicture}
\caption{History of Self-Labeling}
\label{fig:history_selflabeling}
\end{figure}

Self-distillation methods such as BYOL \citep{grill2020bootstrap}, SimSIAM \citep{chen2021exploring}, DINO \citep{caron2021emerging}, along with their variants rely on a simple mechanism: feeding two different views to two encoders, and mapping one to the other by means of a predictor. To prevent the encoders from \textit{collapsing} by predicting a constant for any input, various techniques are employed. A common approach to prevent collapse is to update one of the two encoder weights with a running average of the other encoder's weights. We discuss the particularities of each method.

{\bf BYOL} (bootstrap your own latent) first introduced self-distillation as a means to avoid collapse. BYOL uses two networks along with a predictor to map the outputs of one network to the other.
The network predicting the output is called the \textit{online} or \textit{student} network while the network producing the target is called the \textit{target} or \textit{teacher} network.
Each network receives a different view of the same image formed by image transformations including random resizing, cropping, color jittering, and brightness alterations.
The student network is updated throughout training using gradient descent.
The teacher network is updated with an exponential moving average (EMA) updates of the weights of the online network.
The slow updates induced by exponential moving average creates an asymmetry that is crucial to BYOL's success. 
The loss can be defined as
\begin{equation}
    \mathcal{L}_{\rm BYOL}\left(\theta_{\rm s},\gamma\right)=\mathbb{E}_{(\vx,t_1,t_2)\sim(X,T_1,T_2)}\left[ \left\| \norm(p_{\gamma}(f_{\theta_{\rm s}}(t_1(\vx))))-\norm(f_{\theta_{\rm t}}(t_2(\vx)))\right\|_2^2\right]
\end{equation}
where the two vectors in representation space are automatically $\ell_2$-normalized i.e.
\begin{align}
    \norm(\vv) = \frac{\vv}{\max(\|\vv\|_2+\eps)},
\end{align}
where $\epsilon$ is often set at $1^{-12}$. $f_{\theta_{\rm s}}$ is the online encoder network often denoted as the {\em student} parametrized by $\theta_{\rm s}$, and $p_{\gamma}$ is the predictor network parameterized by $\gamma$. $\vx \sim X$ is the input sampled from the data distribution $X$, and $t_1(\vx), t_2(\vx)$ are two augmented views of $\vx$ where $t_1 \sim T_1 , t_2 \sim T_2$ are two data augmentations. The target network $f_{\theta_{\rm t}}$ is of the same architecture as the student and is updated by EMA with $\xi$ controlling to what degree the target network preserves its history as in
$$\theta_{\rm t}\leftarrow\xi \theta_{\rm t}+(1 -\xi)\theta_{\rm s}$$ 
with initialization $\eta=\theta_{\rm s}$. 

{\bf SimSiam} is aimed at understanding which components in BYOL are most important. SimSiam showed that the EMA was not necessary in practice, even if it led to a small boost in performance. 
This enabled the use of a simplified loss defined by
\begin{equation}
    \mathcal{L}_{\rm SimSIAM}\left(\theta_{\rm s},\gamma\right)=\mathbb{E}_{(\vx,t_1,t_2)}\left[ \| \norm(p_{\gamma}(f_{\theta_{\rm s}}(t_1(\vx))))-\sg(\norm(f_{\theta_{\rm s}}(t_2(\vx))))\|_2^2\right],
\end{equation}
where for clarity we omit the distribution over which $x,t_1,t_2$ are sampled from.
Several works have aimed at understanding how BYOL and SimSiam avoid collapse such as~\cite{tian2021understanding} or~\cite{halvagal2022predictor}, where they found that the asymmetry between the two branches is the key, as well the training dynamics which regularize the variance of the embeddings implicitly.

{\bf DINO} performs a centering of the output of the student network using a running mean (to avoid sensitivity to mini-batch size) and discretize (smoothly) the representations by means of a softmax with a temperate $\tau$ usually taken to be around $0.1$ as in
\begin{equation}
    \mathcal{L}_{\rm DINO}\left(\theta_{\rm s},\gamma\right)=\mathbb{E}_{(\vx,t_1,t_2)}\left[\CE\left(\softmax(f_{\theta_{\rm s}}(t_1(\vx))/\tau),\sg(\softmax(\cent(f_{\theta_{\rm t}}(t_2(\vx)))/\tau))\right)\right],
\end{equation}
where akin to BYOL the teacher again has a moving average of the student network's weights, usually with value $\xi$ following a cosine schedule from $0.996$ to $1$ during training.
The discretization in DINO caused by the softmax can be interepreted as an online clustering mechanism, where the last layer before the softmax contains the clustering prototypes and its weight. As such, the output of the penultimate layer is clustered using the weights of the last layer.

{\bf iBOT} builds on DINO and combines its objective with a masked image modeling objective applied in latent space directly. Here, the target reconstruction is not the image pixels but the same patches embedded through the teacher network.

{\bf DINOv2} further builds on iBOT and improves its performance significantly in both linear and k-NN evaluations by improving the training recipe, the architecture, and by introducing additional regularizers such as KoLeo \citep{sablayrolles2018spreading}. In addition, DINOv2 curates a larger pretraining dataset consisting of 142 million images (further discussion in \Cref{sec:weakly-curated-data}).

Many other methods belong to this self-distillation family. MoCo is another popular method based on building a dictionary look-up that was shown to in some cases to surpass supervised learning on segmentation and object detection benchmarks \cite{he2020momentum}.
Originally the momentum encoder was introduced as a substitute for a queue in contrastive learning \citep{he2020momentum}, which extends the result of \citep{dosovitskiy2014discriminative}.
MoCo's moving average uses a relatively large momentum with a default value of $\xi=0.999$. This higher momentum value works much better than a smaller value of say $\xi=0.9$.
When SimCLR introduced the use of a projector and stronger data-augmentations, MoCoV2 \citep{chen2020improved} followed suite with stronger data-augmentations and a projector head to boost performance.
In a similar spirit, ISD \citep{tejankar2021isd} compares a query distribution to anchors from the student distribution using KL-divergence that relaxes the binary distinction between positive and negative samples.
MSF \citep{koohpayegani2021mean} compares a query's nearest neighbor representation to the student target's representation and then minimize the $\ell_2$  distnace between them with renormalization (akin to cosine similarity maximization).
Another approach, SSCD builds on the contrastive objective to the task of copy detection outperforming copy detection models and other contrastive methods~\citep{pizzi2022selfsupervised}.
Aside from the widespread use of the contrastive objective, many more methods employ similar running average updates as part of their training mechanism. For example, self-distillation \citep{hinton2015distilling,furlanello2018born}, Deep Q Network in reinforcement learning \citep{mnih2013playing}, Mean Teacher in semi-supervised learning \citep{tarvainen2017mean}, and even model average in supervised and generative modeling \citep{jean2014using}.

\subsection[The Canonical Correlation Analysis Family: VICReg/BarlowTwins/SWAV/W-MSE]{The Canonical Correlation Analysis Family:\\ VICReg/BarlowTwins/SWAV/W-MSE}

The SSL canonical correlation analysis family originates with the Canonical Correlation Framework (CCA) \citep{hotelling1992relations}.
The high-level goal of CCA is to infer the relationship between two variables by analyzing their cross-covariance matrices.
Specifically, let $\mX\in \mathbb{R}^{D}$ and $\mY\in\mathbb{R}^{D}$. The CCA framework seeks two transformations $\mU = f_{x}(\mX)$ and $\mV = f_{y}(\mY)$ such that
\begin{gather}
\mathcal{L} = -\sum_{n=1}^{N}\langle \mU_n,\mV_n\rangle,\nonumber\\
 \text{ such that }
\underbrace{\frac{1}{N}\sum_{n=1}^{N}\mU_n=\frac{1}{N}\sum_{n=1}^{N}\mV_n=\mathbf{0}}_{\text{zero-mean representations}},\underbrace{\frac{1}{N}\mU^T\mU=\frac{1}{N}\mV^T\mV=\mI}_{\text{identity covariance representations}},\label{eq:CCA}
\end{gather}
with $d$ (the dimension of the output mappings) such that $d \leq \min (\dim(\mX),\dim(\mY))$. Linear CCA \citep{hotelling1992relations} considers the two mappings to be linear in which case the optimal parameters can be found through the SVD of $\Sigma_{x}^{-\frac{1}{2}} \Sigma_{xy}\Sigma_{y}^{-\frac{1}{2}}$, involving the covariance matrices of $\mX,\mY$ and their cross-covariance. A major advance in the study nonlinear CCA was achieved by \citet{breiman1985estimating} in the univariate output setting, and by \citet{makur2015efficient} in the multivariate output setting, by connecting the solution to \cref{eq:CCA} to the Alternating Conditional Expectation (ACE) method.
\citet{painsky2020nonlinear} study the link between the optimal representation for nonlinear CCA using the Alternating Conditional Expectation proving new theoretical bounds that lead to further refinements of CCA.

These ideas were extended to deep learning in Deep Canonically Correlated Autoencoders (DCCAE) an autoencoder regularized via CCA. 
\citet{hsieh2000nonlinear} and \citet{andrew2013deep} introduce the objective of jointly learning parameters for two networks, $f_1, f_2$, such they their outputs are maximally correlated. The inputs to these networks are two views $X_1$ and $X_2$. Specifically the objective is then to find parameters $\theta_1, \theta_2$ for each network such that
\begin{align}
(\theta_1^*, \theta_2^*) = \text{argmax}_{(\theta_1, \theta_2)} \text{corr}(f_1(X_1; \theta_1), f_2(X_2; \theta_2).
\end{align}

This DCCAE objective was extended to multivariate outputs and arbitrary DDNs in \citet{wang2015deep}.

\begin{figure}
    \centering
    \includegraphics[trim=0.0cm 0.0cm 3.5cm 0.0cm, clip, width=\textwidth]{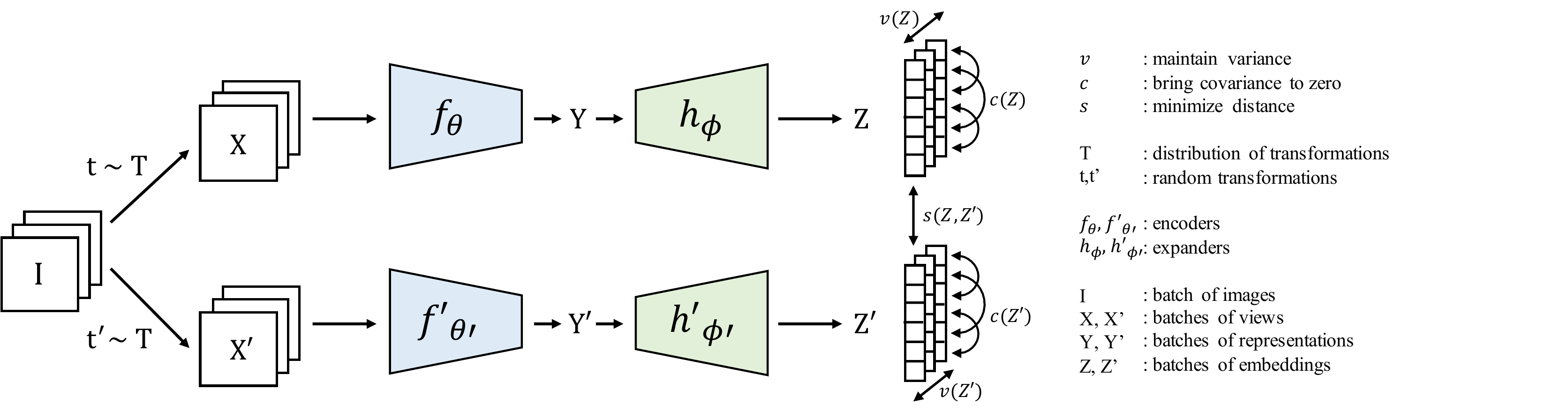}
    \caption{\textbf{VICReg}: penalizes variance, invariance, and co-variance terms to learn representations from unlabeled data.}
    \label{fig:vicreg-diagram}
\end{figure}

From these origins, stems SSL methods such as VICReg \citep{bardes2021vicreg}, Barlow Twins \citep{zbontar2021barlow}, SWAV \citep{caron2020unsupervised}, and W-MSE \citep{ermolov2021whitening}. \textbf{VICReg}, the most recent among these methods, balances three objectives based on co-variance matrices of representations from two views: variance, invariance, co-variance shown in Figure \ref{fig:vicreg-diagram}. Regularizing the variance along each dimension of the representation prevents collapse, the invariance ensures two views are encoded similarly, and the co-variance encourages different dimensions of the representation to capture different features.

\subsection{Masked Image Modeling\label{sec:mim}}

A number of prominent early self-supervised pre-training algorithms for computer vision applied degradations to training images, such as decolorization \citep{zhang2016colorful}, noise \citep{vincent2008extracting}, or shuffling image patches \citep{noroozi2016unsupervised}, and taught models to undo these degradations. Context encoders instead mask out large portions of an image and replace their pixel values with white, teaching an autoencoder to inpaint the white patches \citep{pathak2016context}. This early attempt at masked image modeling does not achieve competitive performance with supervised learning on downstream tasks, and pre-dates vision transformer architectures which modern masked training routines build upon. Subsequently, BERT \citep{devlin2019bert} shook up the natural language processing world by replacing text tokens input to a transformer language model with learnable mask tokens and teaching the model to recover the original text. This paradigm, termed \emph{masked language modeling} (MLM), can also be interpreted as a form of the above strategy, degrading a sample via masking and teaching a model to undo the masking degradation. MLM, along with span-infilling techniques, remains popular as a SSL objective for large language models~\citep{raffel2020exploring,wang_what_2022,tay_unifying_2022}.

We can also similarly mask out portions of an image and teach a model to inpaint them.  This pre-training vision strategy is known as masked image modeling (MIM).  Inspired by BERT, \citet{dosovitskiyimage} exploit the vision transformer architecture by masking out patch tokens and replacing them with learned mask tokens.  They then teach their model to predict pixel values directly, but they find that this pre-training strategy is significantly less effective than supervised pre-training.

\citet{bao2021beit} note that applying the BERT strategy directly to images is difficult because whereas text tokens can only take on a small number of values that can be predicted as a classification problem, image patches can assume considerably more possible values and hence more classes than would be suitable for classification.  Instead, the authors cast MIM as a regression problem, first using an autoencoder to encode image patches as discrete tokens, and then pre-training their transformer to predict the discrete token values for masked tokens.  BEiT achieves significantly improved performance on downstream image classification and semantic segmentation over previous supervised and self-supervised baselines, but its training pipeline is complex since it requires a powerful autoencoder for converting image patches to discrete tokens.

In order to streamline MIM pre-training, two concurrent works \citep{he2022masked, xie2022simmim} propose simplified algorithms, masked autoencoders (MAE) and SimMIM respectively, which directly reconstruct masked image patches rather than discrete image tokens extracted from an encoder as in BEiT.  Moreover, these simplified pre-training strategies achieve superior performance to BEiT on downstream image classification, semantic segmentation, and object detection tasks.  Since then, masked image modeling has achieved competitive performance on a wide variety of vision tasks \citep{zhou2021ibot, woo2023convnext, oquab2023dinov2} and even vision-language representation learning \citep{fang2022eva}. The most successful approaches when using a frozen encoder, iBOT \citep{zhou2021ibot} and DINOV2 \citep{oquab2023dinov2} employ a mix of masked image modeling and more classical approaches such as self-distillation. Howver, their masked image modeling objective reconstructs in latent-space with a teacher network used to provide targets instead of using the original image as the reconstruction target.

Consider that MIM is fundamentally a generative modeling task.  Such models are trained to generate missing image parts conditional on the observed ones. 
Note that BEiT, MAE, and SimMIM are deployed on downstream prediction problems by removing the decoder and replacing it with a prediction head.  However, masked image models can also achieve strong generative modeling \citep{chang2022maskgit}, including text-conditional generation \citep{chang2023muse}.  Compared to autoregressive models for image generation \citep{yuscaling} which generate patches sequentially, MIM-based generative models are significantly more efficient, since they can generate patches in parallel.

In \Cref{subsec:techniques_masked}, we will discuss various techniques harnessed by state-of-the-art masked image modeling systems to achieve such competitive performance.

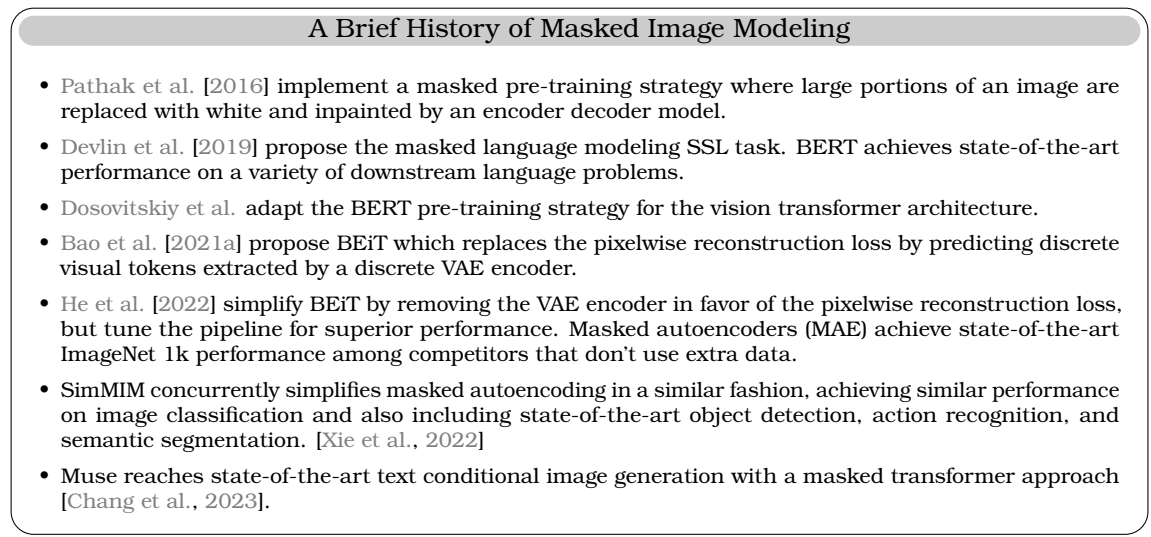
\begin{figure}[h!]
\centering
\begin{tikzpicture}
    \SummaryCard{7}{A Brief History of Masked Image Modeling}{\begin{minipage}{0.95\textwidth}
    \begin{itemize}
    \item \citet{pathak2016context} implement a masked pre-training strategy where large portions of an image are replaced with white and inpainted by an encoder decoder model.
    \item \citet{devlin2019bert} propose the masked language modeling SSL task.  BERT achieves state-of-the-art performance on a variety of downstream language problems.
    \item \citet{dosovitskiyimage} adapt the BERT pre-training strategy for the vision transformer architecture.
    \item \citet{bao2021beit} propose BEiT which replaces the pixelwise reconstruction loss by predicting discrete visual tokens extracted by a discrete VAE encoder.
    \item \citet{he2022masked} simplify BEiT by removing the VAE encoder in favor of the pixelwise reconstruction loss, but tune the pipeline for superior performance.  Masked autoencoders (MAE) achieve state-of-the-art ImageNet 1k performance among competitors that don’t use extra data.
    \item SimMIM concurrently simplifies masked autoencoding in a similar fashion, achieving similar performance on image classification and also including state-of-the-art object detection, action recognition, and semantic segmentation. \citep{xie2022simmim}
    \item Muse reaches state-of-the-art text conditional image generation with a masked transformer approach \citep{chang2023muse}.
    \end{itemize}
    \end{minipage}
    }
\end{tikzpicture}
\caption{A Brief History of Masked Image Modeling}
\label{fig:history_mim}
\end{figure}

\subsection{A Theoretical Unification Of Self-Supervised Learning}

\subsubsection{Theoretical Study of SSL}

Numerous works have attempted to unify various SSL methods. In~\citet{huang2021towards}, Barlow Twins' criterion is shown to be linked to an upper bound of a contrastive loss. 
This suggests a link exists between contrastive and covariance-based methods. This direction was further pursued in~\citet{garrido2022duality}, where a covariance-based and contrastive criterion are shown to be equivalent up to normalization  by deriving the precise gap between the two approaches. These results were further validated empirically as methods were shown to exhibit similar performance and representation properties at ImageNet's scale (1.2 million samples). The similarities among methods was also studied in~\citet{tao2021unigrad} where this unification was tackled from a study of the losses' gradients.

\paragraph{Relationship between Contrastive Learning and Other Objectives.}\label{sec:contrastive}
Initially, InfoNCE was suggested as a variational approximation to the mutual information between two views \citep{aitchison2023infonce, wang2020understanding, oord2018representation}. 
\citet{li2021self} explains the role of InfoNCE in contrastive learning through the lens of the Hilbert-Schmidt Independence Criterion (HSIC), which was used to present a variational lower bound on the mutual information (MI) between different transformations.
\citet{tschannen2020mutual} shows the performance of InfoNCE cannot be explained only in terms of mutual information. Instead other factors such as the feature extractor and formualtion of the mutual information estimator are important and can lead to drastically different performance
 \citep{guo2022tight}. 
Alternative theories suggest that InfoNCE balances alignment of “positive” examples and uniformity of the overall feature representation \citep{wang2022understanding}, or that (under strong assumptions) it can identify the latent structure in a hypothesized data-generating process, akin to nonlinear ICA \citep{khemakhem2020variational}.
In \citet{wang2020understanding}, Theorem 1 shows that contrastive learning with an RBF kernel (an expressive map of features into a higher dimensional space) converges to a uniform distribution on the sphere with matched pairs. ~\citep{tian2022understandinga} shows that contrastive learning with deep linear network is equivalent to Principal Component Analysis (PCA) and ~\citep{tian2023understanding} further analyzes the role played by nonlinearity in the architecture if trained with contrastive loss, showing that nonlinearity leads to many local optima that can host diverse patterns in the training data, while linear networks only allow a single dominant pattern to be learned.   
\citet{hjelm2019learning} introduced Deep InfoMax (DIM), which maximizes the mutual information between the input and output of a deep neural network encoder using local features from the input, an idea that was extended to graphs in \citet{veličković2018deep}.

\def\cL{\mathcal{L}}

\begin{figure}
    \includegraphics[width=.32\textwidth]{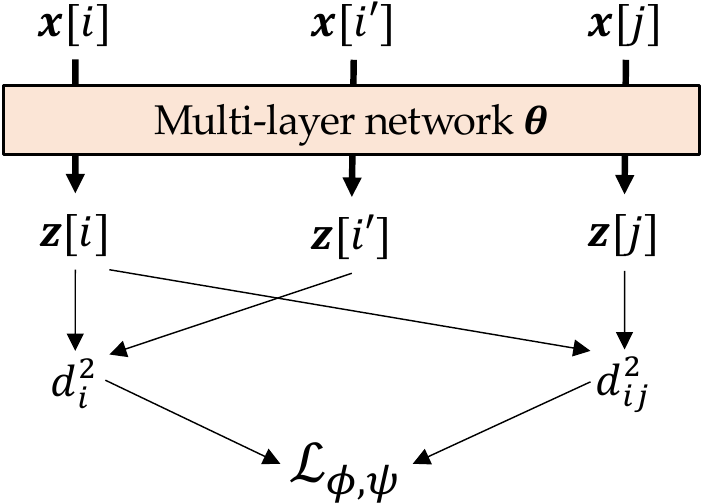}
    \hfill 
    \footnotesize
    \setlength\tabcolsep{2pt}
    \begin{tabular}[b]{l|l|l|}
    Contrastive Loss & $\phi(x)$ & $\psi(x)$ \\
    \hline 
    \hline
    InfoNCE~{\tiny\citep{oord2018representation}} & $\tau\log(\epsilon + x)$ & $e^{x/\tau}$ \\
    MINE~{\tiny\citep{belghazi2018mutual}} & $\log(x)$ & $e^x$ \\
    Triplet~{\tiny\citep{schroff2015facenet}} & $x$ & $[x + \epsilon]_+$ \\ 
    Soft Triplet~{\tiny\citep{tian2020understanding}}  & $\tau\log(1 + x)$ & $e^{x/\tau + \epsilon}$ \\
    N+1 Tuplet~{\tiny\citep{sohn2016improved}}      & $\log(1+x)$ & $e^x$ \\
    Lifted Structured~{\tiny\citep{oh2016deep}} & $[\log(x)]^2_+$ & $e^{x + \epsilon}$ \\
    Modified Triplet~{\tiny Eqn.~10~\citep{coria2020comparison}} & $x$ & $\mathrm{sigmoid}(c x)$ \\ 
    Triplet Contrastive~{\tiny Eqn.~2~\citep{ji2021power}} & linear & linear \\
    \end{tabular}
    \caption{\small Problem Setting. \textbf{Left}: Data points ($i$-th sample $\vx[i]$ and its augmented version $\vx[i']$, $j$-th sample $\vx[j]$) are sent to networks with weights $\vtheta$, to yield outputs $\vz[i]$, $\vz[i']$ and $\vz[j]$. From the outputs $\vz$, we compute pairwise squared distance $d^2_{ij}$ between $\vz[i]$ and $\vz[j]$ and intra-class squared distance $d^2_i$ between $\vz[i]$ and $\vz[i']$ for contrastive learning with a general family of contrastive loss $\cL_{\phi, \psi}$ (Eqn.~\ref{eq:general-loss}). \textbf{Right}: Different existing loss functions corresponds to different monotonous functions $\phi$ and $\psi$. Here $[x]_+ := \max(x, 0)$.}
    \label{tab:loss-funcs}
\end{figure}

\paragraph{Unified contrastive losses.} ~\citet{tian2022understandinga} unified contrastive losses as minimizing a general family of loss functions $\cL_{\phi,\psi}$, where $\phi$ and $\psi$ are monotonously increasing and differentiable scalar functions
\begin{equation}
    \min_{\vtheta} \cL_{\phi,\psi}(\vtheta) = \sum_{i=1}^N \phi\left(\sum_{j\neq i} \psi(\|\vz_i-\vz_{i'}\|_2^2 - \|\vz_i-\vz_{j}\|_2^2)\right). \label{eq:general-loss}
\end{equation}
where $z$ are representations with indices $i$ and $j$ running from $1$ to $N$. With different $\phi$ and $\psi$, Eqn.~\ref{eq:general-loss} covers many loss functions (\Cref{tab:loss-funcs}). In particular, setting $\phi(x) = \tau\log(\epsilon + x)$ and $\psi(x) = \exp(x/\tau)$ gives a generalized version of InfoNCE loss~\citep{oord2018representation}:
\begin{equation}
    \!\!\cL_{nce}\!:=\!-\tau \sum_{i=1}^N\log\frac{e^{-\|\vz_i-\vz_{i'}\|_2^2/\tau}}{\epsilon e^{-\|\vz_i-\vz_{i'}\|_2^2/\tau}+\sum_{j\neq i} e^{-\|\vz_i-\vz_{j}\|_2^2/\tau}} 
\end{equation}
where $\epsilon > 0$ is some constant e.g. $\epsilon = 1$ has been used in \citet{He2020MomentumCF,tian2020contrastive}, $\epsilon = 0$ yields a slight variation of SimCLR \citep{chen2020simple}, the DCL loss \citep{yeh2021decoupled}.

\paragraph{Hard negative sampling.} Negative mining has been thoroughly studied in (deep) metric learning. Recently, some works have focused on putting more weight on hard samples \citep{robinson2020contrastive}. Yet, \citet{kalantidis2020hard,tian2022understandinga} showed that contrastive SSL losses with $\psi=e^{x/\tau}$ already have such mechanisms at the batch level, focusing on {\em hard-negative pairs} without explicit "hard-negative sampling". This means that {\em contrastive losses need large batch sizes to ensure that hard negative samples are observed} which occurs at an additional memory cost.


\label{sec:projector_theory}
\paragraph{Study of the projector.} The projector network, first introduction by~\cite{chen2020simple}, maps the representations into another space where the loss is computed. Despite strong empricial evidence the projector improves performance, few theoretical works attempted to explain its role. \cite{jing2022understanding} study the role of linear projectors in contrastive learning. More precisely, it is argued that the projector prevents dimensional collapse in the representation space and that it only needs to be diagonal and low-rank to do so. Although the proposed method without a projector outperforms SimCLR with a one layer linear projector, for 2- and 3-MLP projectors, performance remains out of reach. \cite{cosentino2022toward} study the interplay of the projector and data augmentations when the augmentations are Lie group transformations, and, as~\citet{mialon2022variance}, provide an explanation on the effect of width and depth of the projector. Further empirical investigations of the role of the projector are presented in section \ref{sec:projector}.

\subsubsection{Dimensional Collapse of Representations}

\begin{figure}[!thbp]
    \centering
    \includegraphics[width=1\textwidth]{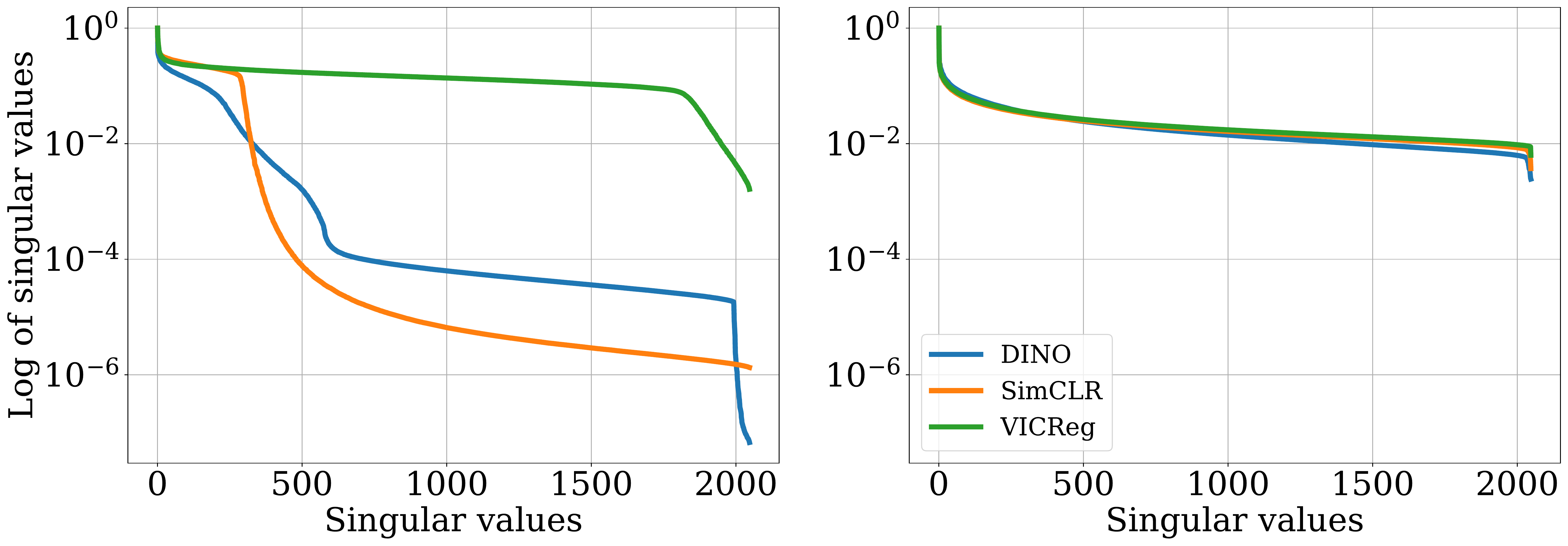}
    \caption{Illustration of dimensional collapse before the projector (Left), and after the projector (Right). Methods suffer from different levels of collapse after the projector; while no such collapse occurs for representations before the projector.}
    \label{fig:collapse}
\end{figure}

While the goal of joint self-supervised methods is to learn meaningful representations, a significant part of the approaches suffer from what is called \textit{dimensional collapse}. 
Dimensional collapse occurs when information encoded across different dimensions of the representation is redundant.
In other words in the output of the projector, the embeddings are rank-deficient, which can be approximated via the singular value spectrum of the embeddings, as illustrated in \Cref{fig:collapse}.

This phenomenon was first illustrated by~\cite{hua2021feature} where the use of a whitening batch normalization helped alleviate collapse. Dimensional collapse was also studied from a theoretical point of view by~\cite{jing2022understanding} with a focus on contrastive methods. Several following works linked dimensional collapse to an impact on performance~\citep{he2022exploring,ghosh2022investigating,li2022understanding,garrido2022rankme}. Some works focused on unsupervised evaluation~\citep{ghosh2022investigating,garrido2022rankme} where dimensional collapse was found to be a good proxy for downstream performance.\\
Different measures of dimensional collapse have been introduced such as the entropy of the singular value distribution~\citep{garrido2022rankme}, the classical rank estimator~\citep{jing2022understanding}, fitting a power law to the singular value distribution~\citep{ghosh2022investigating} or the AUC of the singular value distribution~\citep{li2022understanding}. Nonetheless, all of these measures focus on evaluating the rank of the representations to measure dimensional collapse in the learned representations.

\subsection{Pretraining Data}

\paragraph{Curated (standard)}: The most common practice is to pretrain SSL models on curated datasets such as ImageNet and alternatives such as PASS \cite{asano2021pass}. These datasets tend to generally be class-balanced and contain object-centric images, where the object is prominentely feature often in the center of the photo.

\paragraph{Training with data from the wild}: Even though ImageNet has been the dataset of choice for pretraining, it is definitely not the only option. Its simplicity (object centric, single object, balanced classes) makes it a very good playground but most datasets in the wild are not as clean. If we want to leverage large uncurated datasets for SSL methods need to translate well outside of ImageNet. To this effect some works have explored pretraining on large uncurated datasets~\citep{goyal2021self}, or on datasets that different from ImageNet such as COCO~\citep{el2021large}, or iNaturalist~\citep{uniform_prior}. While these works have shown promising results, ImageNet (or similarly curated dataset) pretraining has remained the norm.

To provide other insights, we pretrained methods on Places205~\citep{zhou2014places} and iNaturalist18~\citep{vanhorni2018naturalist} without changing the augmentations strategy but tuning heavily loss related coefficients. The goal is to see if the setups used on ImageNet transfer well to other datasets. Places205 has the advantage of not being object centric, and iNaturalist of having a power law distribution of classes as well as requiring a lot of fine-grained information. We report our results in \cref{tab:inat-pre}. As we can see most methods are able to achieve similar performance when pretraining either on ImageNet or on the target dataset. This would suggest that the protocol developed on ImageNet can transfer decently, since we noticed that hyperparameters that were optimal on ImageNet also tended to be on different datasets. There is one visbile exception though, SimCLR and MSN perform poorly on iNaturalist18 when pretraining on it directly. While conclusions are impossible to draw precisely here, it would suggest that certain method exhibit more sensitivity on the pretraining dataset than other.

\begin{table}[t!]
    \centering
    \resizebox{0.75\linewidth}{!}{
  \begin{tabular}{lccccccc}
    \toprule
    Target Dataset  & \multicolumn{4}{c}{iNaturalist18} &  \multicolumn{3}{c}{Places205} \\ 
 \cmidrule(lr){2-5} \cmidrule(lr){6-8}  
 Method & VICReg & SimCLR & DINO & MSN & VICReg & SimCLR & DINO\\ 
  \midrule 
 ImageNet pretraining & 38.8 & 39.2 & 46.3 & 40.5 & 52.6 & 51.8 & 54.4\\
 Target dataset pretraining & 37.0 & 28.6 & 41.9 & 29.1 & 53.4 & 51.6 & 57.2 \\
    \bottomrule
  \end{tabular}
  }
    \caption{Comparison of top-1 accuracy on a target dataset by pretraining on ImageNet or on the target dataset directly. We use the same data augmentation strategy as originally developed on ImageNet to study its transferability and heavily tune loss related hyperparameters. Methods were pretained for the same number of iterations on all datasets.}
    \label{tab:inat-pre}
\end{table}

\label{sec:weakly-curated-data}
\paragraph{Weakly-curated training data}:
A successful approach to leverage large uncurated datasets is to perform retrieval in them based on curated data. This means that the dataset will contain images similar to a curated or smaller source dataset such as ImageNet, while being much larger and more diverse. This strategy was used in DINOv2 \citep{oquab2023dinov2} where LVD-142M was built using a wide variety of small and domain specific datasets.
While this does not lead to big performance boosts in classification on ImageNet, it can lead to significant boosts in performance on other tasks such as image retrieval.

\section{A Cook's Guide to Successful SSL Training and Deployment}
\label{sec:practical_matters}

\subsection{Role of Data-Augmentation}
\label{sec:DA}

Many SSL methods, especially joint embedding methods derived from \citet{chen2020simple}, require a way to define positive views from a given image to learn invariances. The proxy used in these SSL methods is to leverage data augmentation to define these invariances. For example, by using different crops of a given images and positive view, the SSL model will be trained to produce a representation that is invariant to these different crops. When using a grayscale operation, or a colorjitter one as positive views, the representation will have to be invariant to the color information. Thus, the deep nature of what is learned by the SSL models is defined by the data augmentation pipeline.  It is worth noting that perfect invariance is not achieved thanks to the projector~\citep{bordes2022guillotine}, which helps improve performance on tasks which are not entirely invariant.
\citet{chen2020simple} study how much influence have specific data augmentations on SimCLR with respect to the performances over ImageNet. They show that simpler data augmentation such as noise aren't beneficial on ImageNet classification downstream. Instead, cropping and multiple color jittering operations lead to competitive results with a supervised baseline. This key element of data-augmentation had also been largely used in the following SSL works \citep{chen2020improved, bardes2021vicreg, zbontar2021barlow} without significant changes. The only variant that is sometimes used is adding smaller crops in addition of bigger crops when learning in-variances. We discuss this use of big and smaller crops, called multi-crop, in the coming subsections. 

However this specific combination of data augmentation was specifically designed to reach good performances on ImageNet. \citet{demo} study the impact of different choice of data augmentation on different downstream tasks and found that even if the addition of ColorJitter seem beneficial for many classification task it might not always be the case on other downstream tasks. Similarly, \citet{ericsson2021selfsupervised} show that different augmentations lead to learning different type of invariances for which some of them are better on some downstream tasks than other. The authors suggest to merge representations learned with different augmentations to improve transferability across a wider range of downstream task. 
There is also an hidden cost when using a complex pipeline of data augmentation: the data preprocessing time which might slow down significantly the training.
 Thus, when the training budget matter, it might be preferable to just use random crop along a grayscale operation when training a SSL model.
 We discuss common approaches for speeding up the training pipeline in \Cref{sec:ffcv}.
 \citet{ni2021close} further show that contrastive learners can benefit from very aggressive data augmentations such as large rotations when explicitly trained not to be invariant to them, as in meta-learning \citep{ni2021data}.

 Another line of work attempts to remove the need for these handcrafted data augmentations. One approach is to use a reconstruction-based objectives such as MAE \citep{he2022masked} which uses  a reconstruction loss in pixel space to avoid the need for defining precise invariances. Another approach is based on a joint-embedding where based on random parts of an images the goal is to predict the representations of the missing parts of the image in the representation space. An example of such method is I-JEPA \citep{assran2023selfsupervised} or Data2Vec2.0 \citep{baevski2022efficient} which use a context part of an image to predict missing small parts of the image. 
Another line of work tries to retain style information about the augmentations to improve downstream performance on tasks requiring style information such as color by predicting style information \citep{xiao2020should,dangovski2021equivariant,gidaris2018unsupervised,scherr2022selfsupervised}. Encoding true equivariance to augmentations (which requires a mapping between embedding) is an active line of work with approaches such as EquiMod\citep{dangovski2021equivariant}, SEN~\citep{park2022learning}, or \citep{marchetti2022equivariant} which also aims at splitting the representations as class and pose. This idea of splitting representations as invariant and equivariant was also explored in SIE~\citep{garrido2023sie} and using Lie group formalism in \cite{ibrahim2022robust}.

\subsubsection{Role of multi-crop}

While works such as MoCo~\citep{meng2021coco} are focused on increasing the number or quality of negative pairs, another direction to improve performance is to increase the number of positives for a given image. Multi-crop, which was introduced with SwAV~\citep{caron2020unsupervised}, tackles this problem by introducing smaller crops ($96\times 96$) on top of the usual two large ones ($224\times 224$). Instead of only comparing the two large crops together, or all pairs of crops, the two large crops are each compared to all other crops (big or small).
As such, if we have 2 large crops and $N$ small crops, the invariance loss is computed $2(N-1)$ times, increasing the positive-pair related signal. The use of smaller crops as well as not comparing all pairs of crops helps reduce the computational cost of these additional crops. While the number of additional crops can vary (10 in Mugs~\citep{zhou2022mugs} compared to 6 in SwAV), it always lead to an icrease in training time and memory usage if used as is.
To mitigate this cost, using $160\times 160$ large crops and 4 $96\times 96$ in SwAV helped mitigate the memory cost and only lead to a training time increase of $25\%$ compared to the classical setting using two crops of size $224\times 224$, while leading to a 4 point performance boost.
As such, multi-crop is a very useful strategy to help boost performance for a marginal additional compute cost. It has thus become almost ubiquitous in recent works~\citep{caron2021emerging,zhou2021ibot,zhou2022mugs,bardes2022vicregl, oquab2023dinov2}. It is worth pointing out that some works have only noticed minor increases in performance~\citep{wang2021solving} where it only lead to a 0.3 point performance increase.

Other approaches have emerged to negate the computational burden of feeding additional crops to the encoder by using nearest-neighbours in embedding space. While with NNCLR~\citep{dwibedi2021little} the matched positive crop is replaced by its nearest-neighbour in latent space, in MSF~\citep{koohpayegani2021mean}, a $k$-NN graph is built in embedding space to provide a similar effect as multi-crop and increase positive-pair related signal. This strategy was further employed in UniVCL~\citep{tang2022unifying} which used augmentation strategies such as edge of node masking in combination with a $k$-NN graph in latent space. All of these approaches show significant performance boosts for a smaller computational cost compared to multi-crop. In MSF, the use of this $k$-NN graph only increases training time by 6\%.

\subsection{Role of the Projector}
\label{sec:projector}

Most SSL with joint embedding methods include a projector (usually 2- or 3-layers MLP with ReLU) after the encoder. The SSL loss is applied to the projector's output, and the projector is usually discarded after training. This crucial component was introduced in SimCLR~\citep{chen2020simple} and, although not responsible for avoiding collapse, allows significant top-1 accuracy gains on ImageNet. For example, in a 100-epochs training, the projector adds around $20 \%$ of top-1 accuracy in SimCLR and VICReg (from around $50 \%$ to $68 \%$ and $48 \%$ to $68 \%$ respectively). \\

\begin{figure}[!thbp]
    \centering
    \includegraphics[width=0.9\textwidth]{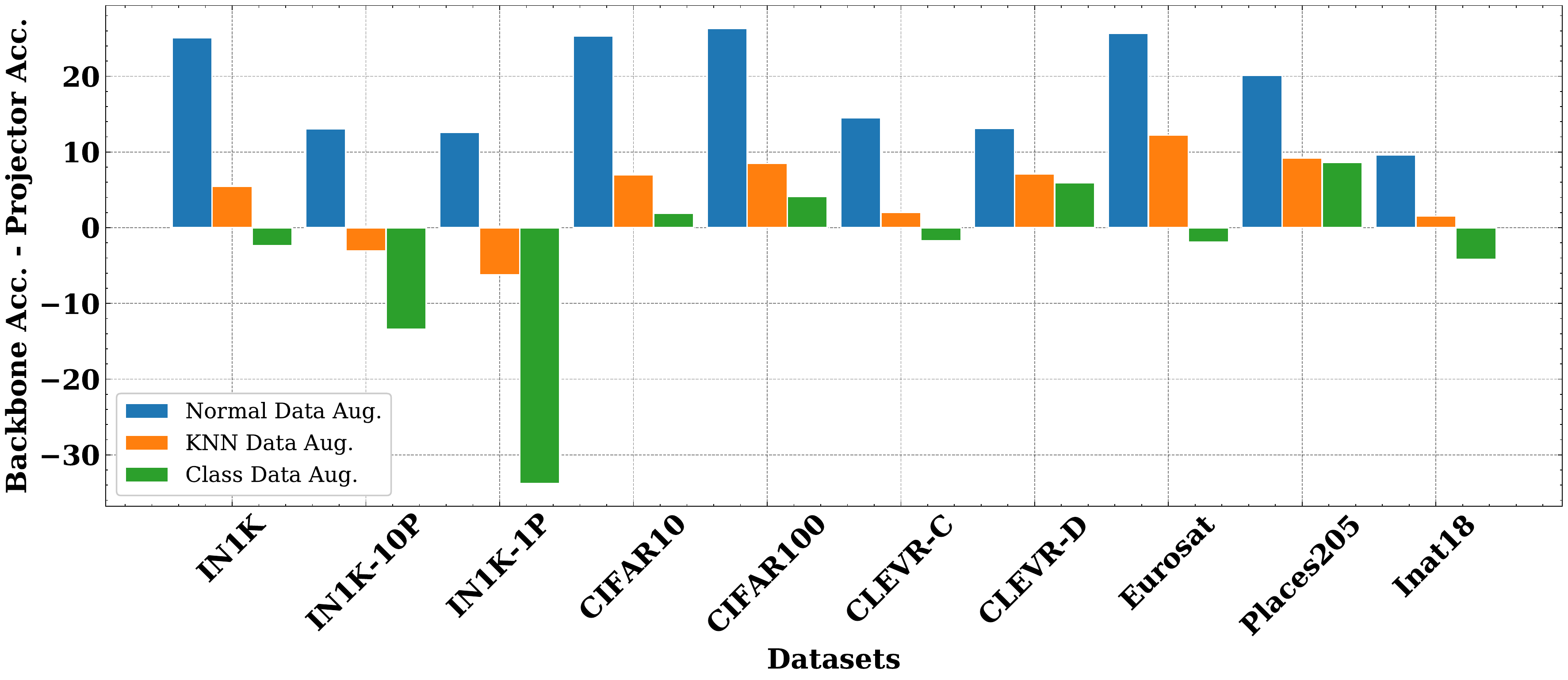}
    \caption{ Figure from~\cite{bordes2022guillotine} that show the accuracy difference between the backbone and projector representation across several downstream tasks. When using the tradition SSL positive pairs (in blue)  the backbone accuracy is always much higher than the projector accuracy. However when using the class label information to define the positive pairs (in green), thus by reducing the misalignement between the training and downstream task, the projector representation lead to higher accuracy than the backbone representation on ImageNet.}
    \label{fig:diff_projector_backbone}
\end{figure}

\citet{bordes2022guillotine} show that adding a projector is not only useful for SSL but is also highly beneficial in a supervised training setting when there is a misalignment between the training and downstream tasks (which was also demonstrated by \citet{sariyildiz2022improving}). In fact, it's well known from \citet{features_transfert} that cutting layers in a trained deep neural network is beneficial when doing transfer learning mostly to avoid the training task's overfitting bias. When looking through the lens of transfer learning, it becomes easy to understand why a projector is needed in SSL since the training task is always different from the downstream task. To bridge the gap between the terms used in the SSL and in the transfer learning literature, \citet{bordes2022guillotine} suggested coining the method of probing intermediate representations or cutting layers as: \textit{Guillotine Regularization} (GR). They also highlight how crucial it is to dissociate GR from the addition of a projector in SSL because the optimal layer on which one should probe the representation might not always be the backbone (but could be an intermediate projector layer as demonstrated in \citet{chen2020big}). Lastly, \citet{bordes2022guillotine} demonstrated that reducing the misalignement between the training and pretext task (by using class label to find the positives pair in contrastive learning) leads to learning a network for which the best linear probe performance on ImageNet are obtained at the last projector layer (instead of the backbone) as shown in Figure \ref{fig:diff_projector_backbone}.

 \begin{table}[h!]
    \centering

  \begin{tabular}{ccll}
    \toprule
     {Projector} & {Oracle} &  {Top-1} & {Top-5}\\ 
  \midrule 
 X & X & 50.1 & 75.8 \\ 
 X & V & 56.4$^{+6.3}$ & 80.2\\
 \midrule 
V & X & 68.9 & 88.2 \\ 
V & V & 69.5$^{+0.6}$ & 88.8 \\
    \bottomrule
  \end{tabular}
    \caption{The projector may handle noise that originates from random data augmentations. Training VICReg without a projector can benefit from filtering semantically inconsistent augmented views using an oracle. With a projector, using an oracle provide only minor gains. Top-1 and Top-5 correspond to linear probing performance on IN-1k.}
    \label{tab:projector_oracle}
\end{table}

\paragraph{Using a projector to handle noisy image augmentations.} The projector may also be necessary to mitigate the noise of data augmentation. As described in \Cref{sec:DA}, SSL methods typically randomly augment input images to generate two different views of the same image. In some cases, enforcing invariance over two very different views might be a very strong constraint that could harm the performance, like when the content of the two views is different. To demonstrate how using the projector can mitigate that, we pretrain VICReg~\citep{bardes2021vicreg} with and without projector using image augmentations that are semantically similar according to an ``oracle'', e.g a ResNet50 pretrained on ImageNet with full supervision. We pretrain for $100$ epochs and include the linear probing results of these experiments in \Cref{tab:projector_oracle}. Without projector and with an oracle, the Top1 performance is $6.3\%$ higher compared to not using an oracle. However, equipped with a projector, using an oracle to remove noisy views only boosts Top1 performance by $0.6\%$. This might imply that the projector has a role in handling inconsistent or noisy augmented views during the SSL training process.

\paragraph{Influence of the projector's output dimension.}
Similarly to how large batch sizes were seen as a requirement for contrastive methods, a large output dimension of the projector was seen as a requirement for covariance based methods. This is illustrated by figure 4 in~\cite{zbontar2021barlow}, and table 12 in~\cite{bardes2021vicreg}, where drops of up to $15\%$ in top-1 on ImageNet can be observed.
As pointed out in~\cite{garrido2022duality} this was due to the projector's intermediate layers scaling with the output dimension as well as loss weights that needed to be scaled as well. By tuning these parameters, VICReg's top-1 accuracy increases from $55.9\%$ to $65.1\%$ with 256 dimensional embeddings. The peak performance is also achieved at 1024 dimensions and plateaus afterwards. 
While VICReg stays more sensitive to the output dimension of the projector than SimCLR, it is significantly more robust than originally thought and very large output dimensions are not a requirement. Comparable results should be achievable for Barlow Twins due to the similarities between the two methods.

\begin{figure}[!thbp]
    \centering
    \includegraphics[width=1\textwidth]{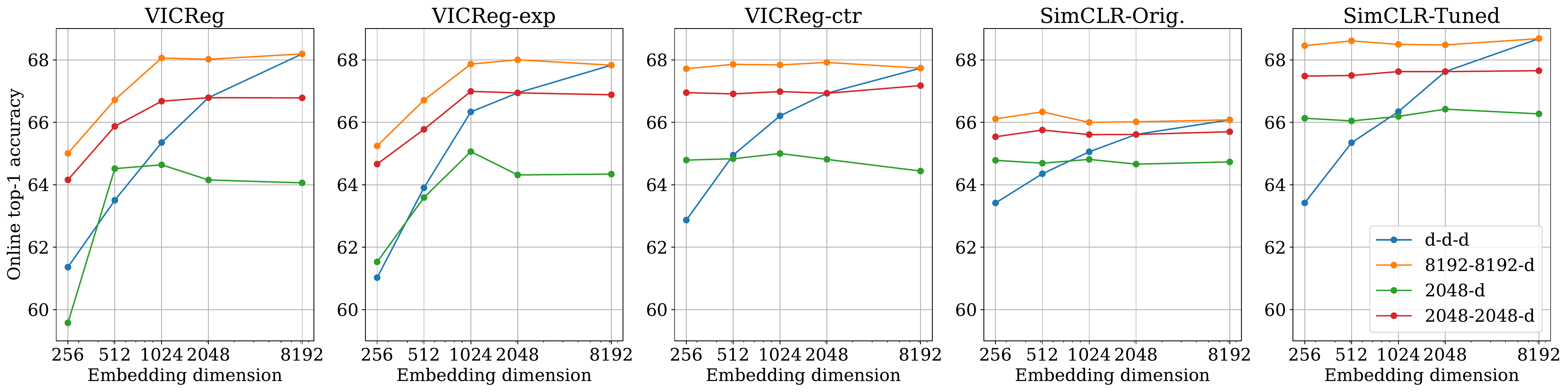}
    \caption{Impact of different projector architectures and output dimension on popular methods.$x-y-z$ denotes a MLP with layers of output dimension $x$,$y$ ad $z$ respectively. From~\cite{garrido2022duality}.}
    \label{fig:projector_architecture}
\end{figure}

\begin{figure}[!th]
    \centering
     \begin{subfigure}{0.45\textwidth}
         \centering
         \includegraphics[scale=0.35]{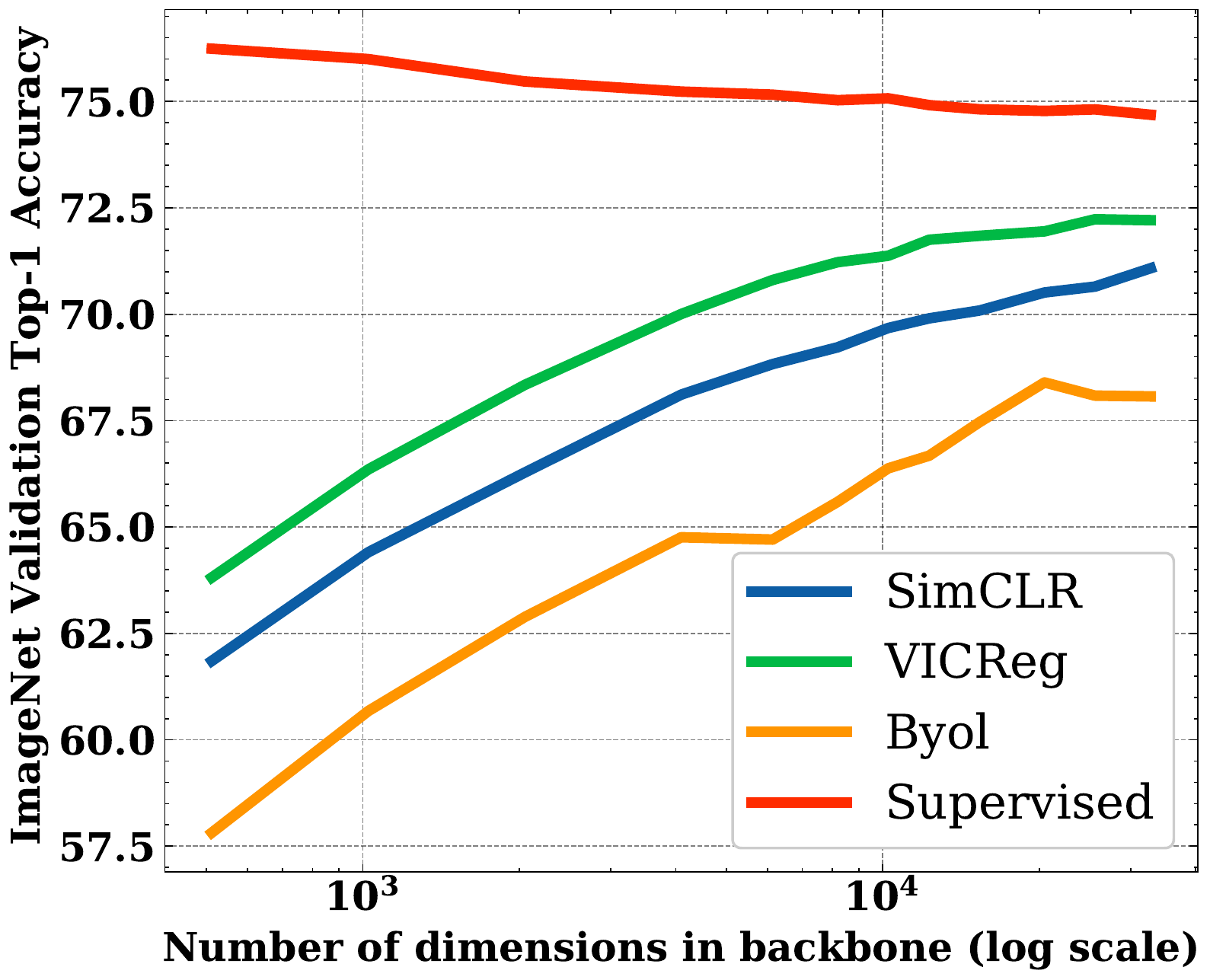}
         \caption{}
         \label{fig:sup_vs_ssl_backbone}
     \end{subfigure}%
    \begin{subfigure}{0.45\textwidth}
         \centering
         \includegraphics[scale=0.3]{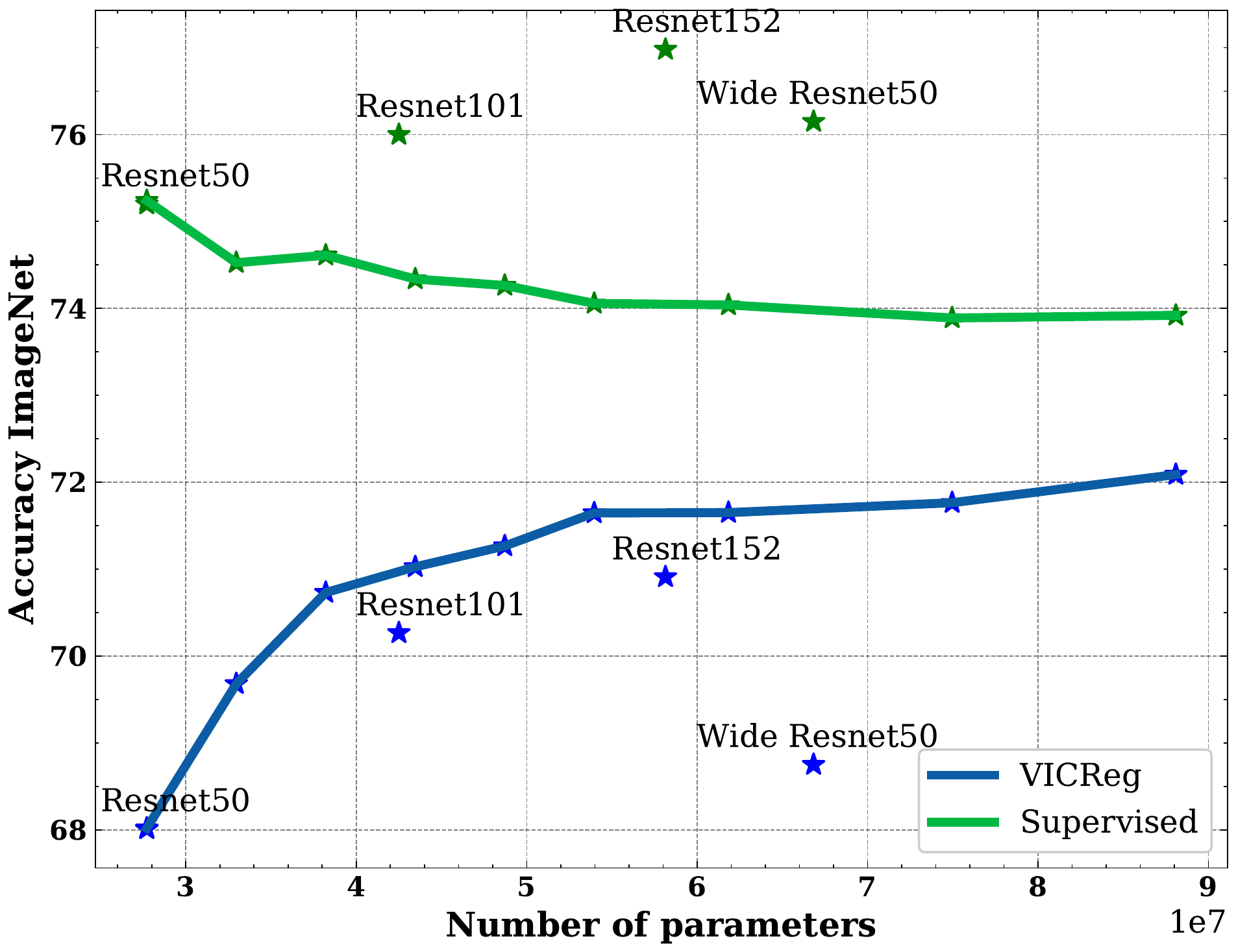}
         \caption{}
         \label{fig:params_vs_acc}
     \end{subfigure}
    \caption{Figures from \citet{bordes2023surprisingly} . a) ImageNet accuracy of various SSL methods with respect to the backbone output dimension. b) ImageNet accuracy with respect to the number of parameters. The dots on the blue and green line are models trained for different backbone output dimension. }
    \label{fig:collapse}
\end{figure}

\paragraph{Influence of the backbone's output dimension.}
Recent works also investigated the effect of the backbone dimension. \citet{dubois_improving_2022} observed that larger backbone representations lead to better linear probe performance when using CISSL. \citet{bordes2023surprisingly} investigated more deeply the impact of the backbone dimension across common SSL methods like VICReg, SimCLR or BYOL. They show that traditional supervised methods decline in performance when the dimension of the backbone is increased. On the other hand, SSL methods highly benefit from wider backbone representations as shown in \Cref{fig:sup_vs_ssl_backbone}. In fact, it is much more beneficial in SSL to increase the backbone size when training a ResNet than increasing the width or depth of the ResNet as illustrated in \Cref{fig:params_vs_acc}. This observation highlights that the current architectures used in SSL, which are often the same as the those used in supervised training, might not be optimal. 

\paragraph{Properties of the representation induced by the projector.} \citet{mialon2022variance} argue that the projector enforces pairwise independence of the features in the representation and provide a demonstration for random projectors in the context of VICReg, BarlowTwins and W-MSE~\citep{bardes2021vicreg, zbontar2021barlow, ermolov2021whitening}. In particular, higher degrees of independence are reached with wider projectors. Pairwise independence, or a soft notion thereof, can be more appropriate to learn unsupervised representations from “real world” datasets such as ImageNet than mutual independence~\citep{li2019learning}. Alternatively, seeking alternative SSL regularization to VCReg is needed if mutual independence is sought for. The optimization dynamics resulting from applying VCReg (the anti-collapse term in VICReg) at the projector's output is also worth noting: minimizing VCReg with respect to the projector parameters is not necessary, and VCReg is rather optimized with respect to the encoder parameters. Whether this analysis fully extends to other SSL methods is an open question.

\paragraph{Training a SSL without a projector.} \citet{jing2022understanding} proposes DirectCLR, which shows that regularizing the representation in DirectCLR by applying the InfoNCE SimCLR objective on sub-vectors of the representation without a trainable projector is sufficient to outperform SimCLR with a linear projector in terms of ImagNet top-1 accuracy.

\subsection{The Uniform Prior in SSL or the Failure of SSL on Unbalanced Data}
 Despite their recent successes, there is an important limitation of SSL methods: poor performance on unbalanced datasets. Since real world data is imbalanced, such a limitation is an important factor that made the use of SSL methods on vast amount of uncurated data challenging. \citet{uniform_prior} explains such a limitation by the use of an hidden uniform prior that is common to many SSL methods. By distributing the data uniformly in the representation space, SSL methods learn to find the most discriminative features in a given mini batch. 
 When data is uniformly distributed across classes labels, the most discriminative features that the model will learn will be class specific. 
 However, when using imbalanced data, the most discriminative features inside the mini batch might not be the class anymore but more low level information which decrease the performances on downstream classification tasks. To alleviate this issue, \citet{uniform_prior} introduce the use of an additional regularization term on the SSL method MSN \citep{msn} to change the distribution of the SSL clustering.

\subsection{Teacher-Student Architecture Specific Tricks}
\subsubsection{Role of the Moving Average Teacher}

While the original BYOL method is based on exponential moving average (EMA) updates of the weights for the target (teacher) network, it was later confirmed that EMA is not necessary (i.e., the online and target networks can be identical). This is also confirmed with SimSiam \citep{chen2021exploring}, as long as the predictor is updated more often or has larger learning rate compared to the backbone.
In the case of DQN, the target network with EMA is shown to remove bias \citet{fan2020theoretical} and \citet{piche2021beyond} showed that the EMA could be removed from the target network by using the correct regularizer.
For BYOL, a stop gradient of the online network, meaning the decay rate is 0 for the target network, collapses as shown in Table 5 of~\cite{grill2020bootstrap}.
\citet{pham2022pros} shows the idea of exponential moving averages provide training stability that can even be used in non student-teacher frameworks such as SimCLR. Specifically, they show applying EMA updates to the projector of SimCLR can boost performance.~\citet{wang2022importance} shows that training could also benefit from other kinds of asymmetries in the teacher-student setting (e.g., stronger augmentation on the student side).  

\subsubsection{Role of the Predictor in Self-Labeling SSL}
\label{sec:predictor}

The predictor network plays a central role in BYOL's success by predicting the representation of the teacher network from the student networks' representation. 
\citet{shi2020run} shows removing the predictor leads to a performance drop from 68\% to 21\% top-1 accuracy on ImageNet (compared to the original two-layer MLP predictor in BYOL).
In Figure 1 of \citet{shi2020run}, they demonstrate even a linear predictor leads to good performance and can recover from poor initialization in 10-20 epochs of training.
For SimSiam, Table 1 of \citet{chen2021exploring} shows removing the predictor in SimSiam also leads to collapse with a top-1 accuracy of < 1\% on ImageNet.
\citet{tian2021understanding}, whose implementation can be found\footnote{\url{https://github.com/facebookresearch/luckmatters/tree/main/ssl}}, proves that in the presence of the predictor, the training dynamics of BYOL and SimSiam contains nontrivial stable fixed points, and thus avoid being trapped into trivial solutions during training, even if these trivial solutions are global optimal. It further proposed a contrastive method, DirectPred, that directly sets the predictor via eigenvalue decomposition during training and leads to comparable performance in ImageNet. Its follow-up work (DirectSet~\citet{wang2021towards}) further removes the overhead of eigenvalue decomposition.  

\subsection{Role of Standard Hyper-Parameters}
A common issue in SSL research is that each method has different configurations of hyper-parameters. Hence comparisons directly between different SSL methods or models is often challenging. In this section, we present and describe the impact of each hyper-parameters to help SSL practitioners identify which are most important depending of their setup.

\begin{tikzpicture}
    \Quote{When debugging don’t trust the value of the loss, and first play with loss hyper-parameters (not DA/optimizer)}{Adrien, Quentin}
\end{tikzpicture}

\subsubsection{Role of Mini-Batch Size}

It was originally thought that contrastive methods such as SimCLR or MoCo require large batch sizes or memory banks to work. This turns out to be misleading as both methods can be made to work at small batch sizes. A square root scaling of the learning rate was discussed in the appendix of~\cite{chen2020simple} which already gave a significant increase in performance of up to 5 points in top-1 accuracy on ImageNet for a 100 epochs training. Similarly, \citet{demo} investigated the impact of the learning rate with small batch sizes and found how one can train SimCLR on ImageNet using a single gpu without an important drop in performances.
Furthermore, some works such as DCL~\citep{yeh2021decoupled} show that you can reach top performance with a batch size of 256 or more for SimCLR, and a queue size of only 256 or more for MoCo, by simply removing the positive pair from the denominator of the softmax and with more careful hyperparameter tuning. Similarly, it was shown by~\cite{zhang2022dual} that by decomposing the dictionary in MoCo and by using different temperatures for the positive and negative pairs it is possible to increase the robustness to the dictionary dimension. 

\begin{tikzpicture}
    \Quote{Set batch size to maximum that fits on your GPU}{Adrien}
\end{tikzpicture}

\subsubsection{Role of Learning Rate (Schedulers) and Optimizers}

Here we overview typical standard settings for learning rate schedulers and optimizers across methods.
To determine the learning rate, methods often scale a base learning rate based on the batch size according to the heuristic by \cite{goyal2017accurate}: learning rate = $\frac{\text{batch size}}{256} * \text{base learning rate}$. 
For ImageNet pretraining, VICREg, Barlow Twins, BYOL, and SimCLR use a base learning rate of $0.2-0.3$ with the LARS optimizer \citep{you2017large}. 
Additionally for some methods such as Barlow twins, a much smaller learning rate (0.0048) is used to update the bias terms and batch norm parameters.
Other methods such as MAE, DINO, and iBot use the AdamW optimizer \citep{loshchilov2017decoupled} with a smaller base learning rate of $1e-5-5e-4$. For a discussion of weight decay see \Cref{subsec:weightdecay}. 
The most common training schedule involves a warmup period, usually 10 epochs, where the learning rate is linearly increased to its base value. After the warmup period, most methods use cosine decay.

\begin{tikzpicture}
    \Quote{AdamW/LARS with the standard linear warmup/cosine annealing learning rate schedule is a safe choice}{Adrien, Quentin}
\end{tikzpicture}

\subsubsection{Role of Weight-Decay}
\label{subsec:weightdecay}
Weight-decay is an important component of backprogagation for many SSL methods. 
Table 15 in BYOL \citep{grill2020bootstrap} indicates that no weight decay may lead to unstable results. A recent blogpost\footnote{\url{https://generallyintelligent.ai/blog/2020-08-24-understanding-self-supervised-contrastive-learning/}} also mentions using weight decay leads to stable learning in BYOL.
In Figure 4 of \citet{tian2020understanding} the effect of weight decay is explained in terms of its effect on memory of the initial conditions. The hypothesis is that weight decay allows the online network and predictor to better model invariance to augmentations regardless of the initial condition.
For further reading, \citet{zhang2022does} provides a good review of SimSIAM collapse understanding and \citet{shi2020run} does the same for BYOL.

\subsubsection{Vision Transformers Considerations}

Training Vision Transformers (ViT)~\citep{dosovitskiyimage} requires special care. They are more prone to collapse and instability, and are more sensitive to the setting of hyper-parameters~\citep{touvron2021training}.

\textbf{Batch size.}
\citep{chen2021empirical} found that large batch (e.g., 4096) training for joint-embedding ViT SSL methods can be unstable. This instability does not reflect as a large drop in the final accuracy, but appears as drops in kNN probe accuracy during training when the $L_\infty-norm$ of the gradient spikes. Using a random (versus a learned) patch projection layer to embed pixel patches into input tokens for ViT stabilizes training for MoCo-V3, SimCLR, and BYOL and also improves the final accuracy. A learning rate warm-up period of 10k iterations \citep{goyal2017accurate, dosovitskiyimage} also improves training stability. On the other hand, \cite{caron2021emerging} noted a drop in final k-NN accuracy when training with very small batch sizes (128). So, a batch size of 1024 or 2048 seems to be the sweet spot for SSL pre-training of ViTs. 

While the ViT architecture does not have any BatchNorm layers, training a MoCo-V3 model with BN layers in the projector heads improved the linear probing accuracy of the ViT \citep{chen2021empirical}. Note that for joint embedding methods, batching can be done either together for all samples and crops in one batch, or separately for each batch of crops. SimCLR adopts the former, while BYOL and MoCo-V3 adopt the latter.

\textbf{Patch size.}
\citep{caron2021emerging} found that training with smaller patch sizes ($5\times5$, or $8\times8$ instead of $16\times16$) leads to improved linear probing accuracy on DINO ViT pre-training. Note that while increasing patch sizes leads to a reduction in running time, it also increases memory usage (which makes it hard to train on patches smaller than $8\times8$).

\textbf{Stochastic depth}~\citep{huang2016deep} originated from NLP and was subsequently used in vision models~\citep{touvron2021going} to train deeper models. It randomly drops blocks of the ViT as a regularization. The per-layer drop-rate may depend linearly on the layer depth or uniformly as suggested in recent works~\citep{touvron2021going}. It has huge importance when training larger models (ViT-L, ViT-H, etc.). For instance \cite{touvron2022deit} use $0.5$ drop path rate for ViT-H models. Conversely, when training smaller models like ViT-B, such regularization usually hurts the performance~\citep{steiner2021train}.

\textbf{LayerDecay}~\citep{clark2020electra} decreases the learning rate geometrically along the layers. Put differently, the last layer is not affected, while the first has very small learning rate. In SSL vision models, LayerDecay increases performance when fine-tuning on downstream tasks~\citep{beit, zhou2021ibot, he2022masked}. Depending on the model size, the parameter is set between $0.65$ to $0.85$ -- larger models usually need higher values because there are more layers. 
The underlying principle is that SSL builds strong model backbones, therefore we only need to fine-tune the shallowest layers.

\textbf{LayerScale}~\citep{touvron2021going} is a per-channel multiplication of the vector produced by each residual block of the transformer. It increases the stability of the optimization and permits deeper ViT (larger than ViT-B).

\textbf{ \texttt{[cls]} token.}
When it is not explicitly needed by the method, using the average of the patch tokens instead of the class token saves memory without much change on the accuracies of the network~\citep{zhai2022scaling}.

\subsection{Techniques for High Performance Masked Image Modeling}
\label{subsec:techniques_masked}

While there are several approaches to masked pretraining, the state-of-the-art systems that employ them tend to pair MIM with other techniques. For example, the ConvNextV2 architecture, which was state of the art on ImageNet (for models trained with only public data) when released, employs MAE pretraining \citep{woo2023convnext}. Interestingly, the authors point out that simply pretraining a ConvNextV2 with the MAE framework is subpar. They propose adding a novel normalization layer, called ``global response normalization,'' that proves vital to reaching state-of-the-art results \citep{woo2023convnext}.

In other works that claim state-of-the-art performance on image classification and semantic segmentation, MIM pretraining is paired with distillation. 
While some MIM routines involve reconstructing the masked portion of the input in pixel space, another option is to use a teacher network to generate target representations of the unmasked image. 
\citet{zhou2021ibot} propose iBOT, which uses ViTs for both the teacher and the student in distillation-based MIM and outperforms prior methods on ImageNet classification. 
Subsequently, \citet{liu2022exploring} propose dBOT, an updated distillation-based MIM approach which also achieves state-of-the-art results on image classification and semantic segmentation. 
A major finding in their work is that the choice of the teacher model does not have to be chosen carefully if the distillation is done in stages. This is where the teacher is updated periodically to match the student's weights and the student is reinitialized.
\citet{oquab2023dinov2} employ similar distillations to train smaller models from a ViT-g teacher with much better performance than training from scratch.
This line of work highlights that pairing distillation with MIM is extremely effective.

For object detectors that utilize MIM to outperform prior work, techniques that allow MIM to work with recent and high performing pyramid ViTs like Swin are critical.  
Since pyramid ViTs collapse patches, random masking can leave some local windows with no information. \citet{li2022uniform} propose an approach to masking that accounts for the hierarchical structure of these models called ``uniform masking.'' 
They constrain the masking to hide equal amounts of information in each local window ensuring that each has some information intact.
This technique helps self-supervised models (on ImageNet1K) outperform supervised models (even on ImageNet22K) on object detection benchmarks \citet{li2022uniform}.

\subsection{Evaluating Your SSL Models}

\subsubsection{Evaluation with labels}

Self-supervised pre-training is mainly evaluated on image classification, since it has been at the core of computer vision for decades.
The three main common protocols are referred to as $k$-nearest neighbors (KNN), linear and full fine-tuning evaluations (ranked by order of complexity).
They are offline evaluations, meaning that they are done independently of the self-supervised training procedure, conversely to online evaluation, which are performed during training.
While online evaluation can provide a useful signal of downstream performance, because it's optimized alongside the varying self-supervised learning objective, it can be misleading.
In addition, to these procedures which require labels for the downstream task,
more recently, RankMe~\citep{garrido2022rankme} has appeared as a viable alternative to costly evaluations, and is used as an oracle to final accuracy without having to do any training.

\paragraph{KNN}

is one of the best known algorithms of machine learning and has been extensively used throughout the fields.
With regards to image classification, a KNN classifier determines the label of a data point from the labels of its neighbors.

Formally speaking, the model is first used to extract frozen features $\mathcal{X} = x_1, ..., x_n$ (often $l_2$-normalized), from all the images in the training dataset. 
To classify a new image, we extract its feature representation $x'$, and retrieve its $k$ nearest-neighbors.
They are the $k$ vectors of the training set $\mathcal X$ that have highest cosine-similarity with $x'$.
Then, the vanilla approach applies a majority voting scheme: every neighbor counts $+1$ in its corresponding label, and we choose the label with most votes at the end.
More sophisticated approaches use a weigthed voting scheme. 
Instead of counting $+1$ in its corresponding label, every neighbor counts a weight $w = f(x^Tx')$, for instance DINO implemenation employs $w = e^{x^Tx'/T}$ ~\citep{caron2021emerging}. 
This allows to account for imbalanced training set, not i.i.d. features, and usually gives more accurate results, at the cost of introducing an additional hyperparameter $T$.

K-NN classifiers have the great advantage of not relying on many hyperparameters, being fast and light to deploy, without requiring any domain adaptation. 

\paragraph{Linear} 
In the context of SSL evaluation, training a linear classifier on top of pre-trained feature representations, a.k.a. linear-probing evaluation, was introduced by~\cite{zhang2016colorful, zhang2017split}.
It is the most popular protocol for several reasons: it achieves high-accuracy, its performance heavily rely on the quality of the representation since their discriminative power is low, it imitates how the features can be used in practice, and last but not least, it is not very computationally expensive.

\begin{figure}
    \centering
    \includegraphics[scale=0.4]{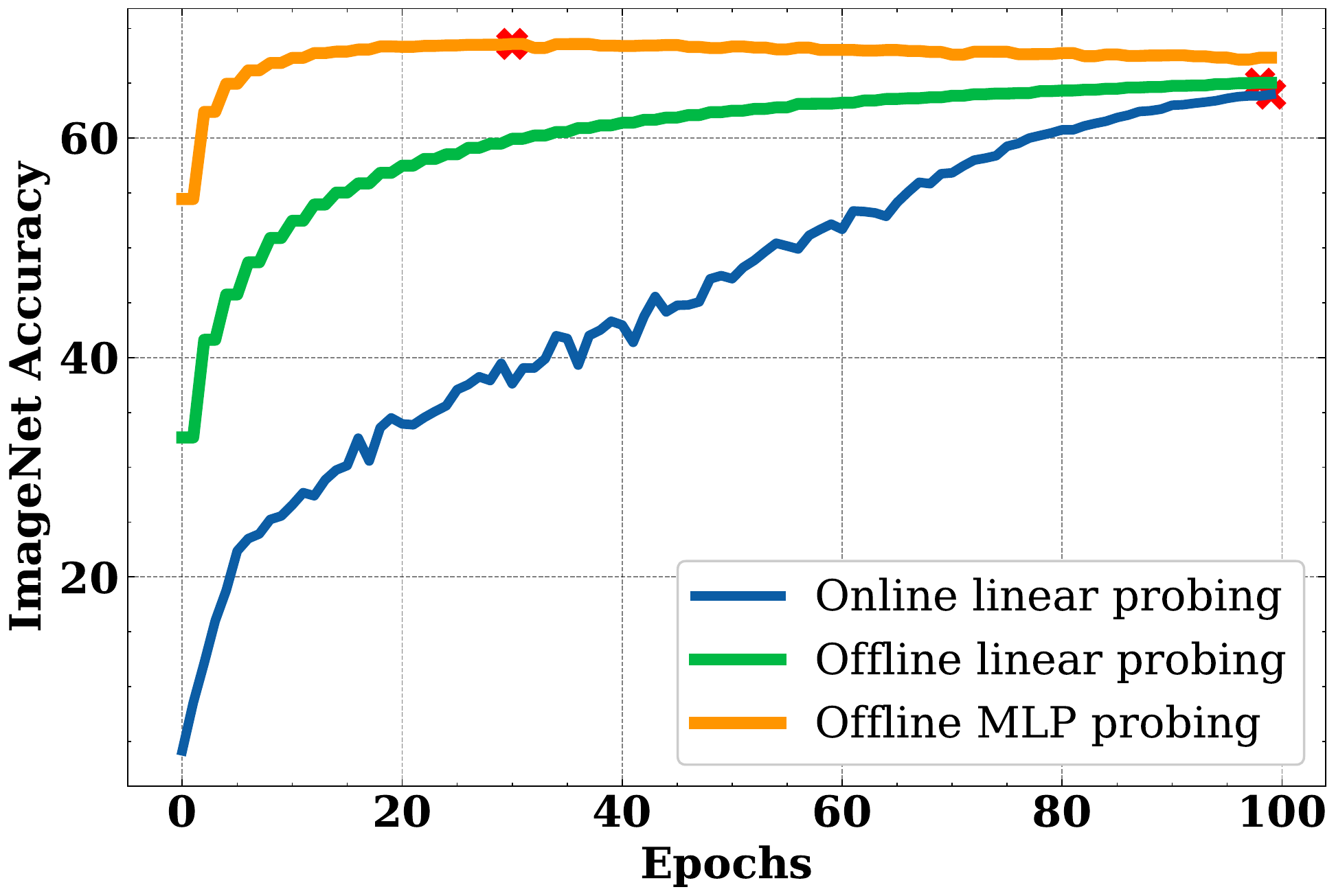}
    \caption{Figure from \citet{demo}. Depiction of the classifier probe trained to predict the Imagenet-1k labels from the output of a Resnet50 backbone during SimCLR training ({\bf online}) and post-training ({\bf offline}) using a linear or MLP classifiers. The cross in red corresponds to the best accuracy. In the offline setting no data-augmentation is employed. We observe clearly that (i) when employing an MLP only a few epochs are needed and regularization or early-stopping should be employed, however, in the popular linear case, we clearly see that there is limited differences between the online and offline performances, and that over-fitting never occurs during either of the training cases. 
    }
    \label{fig:linear_vs_mlp}
\end{figure}

Most of the time, it is done simply by appending a linear layer at the end of the frozen backbone, and optimizing its parameters for a few epochs (around $100$). 
Sometimes, as introduced by \cite{beit}, we can benefit from the fact that the linear evaluation is lightweight and evaluate multiple linear heads at the same time, to test many hyper-parameters at the same time (learning rate, averaging features or using a class-token for ViT-like architectures, number of features, etc.).
A linear probe can also be trained online, by simply cutting gradient from the representations. 
Though only an approximation, an online linear probe is extremely cheap as it reuses the computations for the SSL pretraining, and gives a good indication of downstream performance, as shown in \Cref{fig:linear_vs_mlp}.

\paragraph{MLP} Instead of a simple linear probing, a multi-layer perceptron (with two or three layers) could also be used to extract which information is learned in a SSL model. 
Non linear evaluation is rarely present in work around SSL, however it is needed when the learned features are not linearly separable, and when it is too difficult to extract information present in features with a linear model. In fact, comparing results with a linear and a non linear probe, can give us some ideas about how well structured a representation is. \citet{demo} present some results that compare different evaluation regime using a linear or a non linear probe. In \Cref{fig:linear_vs_mlp}, one can observe that it's possible to get some gain in accuracy when using a multilater layer perceptron instead of a linear probe. However, the main issue with adding capacity into the probe is one related to overfitting: the best MLP head might not be the ones you get after 100 epochs, as showed in \Cref{fig:linear_vs_mlp}.

\begin{tikzpicture}
    \Quote{If labels are available, evaluate your model using an online linear probe}{Adrien, Quentin, Florian}
\end{tikzpicture}

\paragraph{Full Fine-tuning} 
The Masked Auto-encoders (MAE) paper~\citep{he2022masked} re-introduced fine-tuning as the main evaluation metrics.
The main arguments are that linear-probing is uncorrelated with fine-tuning and transfer learning performances, and that small MLP heads do not evaluate the strength of the method to create strong but non-linear features. 
The majority of works that followed~\citep{beit, zhou2021ibot, dong2021peco} focused on this type of evaluation (and sometimes do not report linear/MLP results).
It has been shown that contrasting methods show inferior performance than masked image modeling with regards to fine-tuning, because they are less ``optimization friendly''~\citep{wei2022contrastive} - which explains the overall interest over MIM.
It is by far the most computationally expensive of the evaluation methods, since it needs to re-train the whole network.
The most common benchmark on ImageNet runs the optimization over $100$ epochs for ViT smaller than base, and for $50$ epochs for larger models~\citep{he2022masked}.
Other works~\citep{beit, peng2022beit, wang2022image} first fine-tune on ImageNet-21k for $60$ epochs, and further fine-tune on ImageNet-1k, which represents between $1/5$ to $2$ times the cost of the pre-training phase.

\subsubsection{Evaluation without labels}

As we just discussed, most evaluations rely on the use of labels and training an auxiliary model. This can make evaluations expensive and sensitive to hyperparameters or their optimizations. To help alleviate these issues multiple methods have been proposed to evaluate or help tune hyperparameters of methods without relying on labels.
Using a pretext-task such as rotation prediction can facilitate performance evaluation without labels, as demonstrated in~\cite{reed2021selfaugment} for data augmentation policy selection. However, a drawback of this approach is the requirement for training the classifier for the pretext-task and the assumption that rotations were not part of the pretraining augmentations, or the model would be invariant to it. The eigenspectrum of representations is used in conjunction with the loss value to evaluate performance in ~\cite{li2022understanding}. While a correlation with performance is shown, it requires training a performance classifier with the rank and loss value, making it hard to use for unsupervised evaluation.
In ~\cite{agrawal2022alphareq} $\alpha$-ReQ is introduced to evaluate methods by looking at the eigenspectrum decay of representations before the projector.

 \begin{table}[t!]
    \centering
    \resizebox{0.75\linewidth}{!}{
  \begin{tabular}{llccccccccc}
    \toprule
     \multirow{2}{*}{Dataset} & \multirow{2}{*}{Method}  & \multicolumn{2}{c}{VICReg} &  \multicolumn{1}{c}{SimCLR} &  \multicolumn{2}{c}{DINO} \\ 
 \cmidrule(lr){3-4} \cmidrule(lr){5-5} \cmidrule(lr){6-7}&  & cov. & inv. & temp. & t-temp. & s-temp.\\ 
 \midrule 
\multirow{3}{*}{ImageNet} & \textcolor{gray}{ImageNet Oracle} & \textcolor{gray}{68.2} & \textcolor{gray}{68.2} & \textcolor{gray}{68.5} & \textcolor{gray}{72.3} & \textcolor{gray}{72.4}\\ 
 & $\alpha$-ReQ & \textbf{67.9} & 67.5 & 63.5  & 71.7 & 66.2\\ 
 & RankMe & 67.8 &\textbf{ 67.9} & \textbf{67.1}  & \textbf{72.2} & \textbf{72.4}\\ 
  \midrule 
\multirow{3}{*}{OOD} & \textcolor{gray}{ImageNet Oracle} &  \textcolor{gray}{68.7} & \textcolor{gray}{68.7} & \textcolor{gray}{68.7} & \textcolor{gray}{71.9} & \textcolor{gray}{72.5}\\ 
 & $\alpha$-ReQ  & \textbf{68.1} & 67.8  & 65.1 & \textbf{71.8} & 68.5\\ 
 & RankMe  & 67.7 & \textbf{68.3} & \textbf{67.6} & \textbf{71.8} & \textbf{72.5}\\  
    \bottomrule
  \end{tabular}
  }
    \caption{Hyperparameter selection using the common supervised linear probe strategy (ImageNet oracle), RankMe and $\alpha$-ReQ. OOD indicates the average performance over iNaturalist18, Places 205, Sun397, EuroSat, StanfordCars, CIFAR-10, CIFAR-100, Pascal VOC2007, CLEVR-cnt and FOOD101. Without any label, optimization or parameters, RankMe recovers most of the performance obtained by using ImageNet validation set, highlighting its strength as a hyperparameter selection tool. From~\cite{garrido2022rankme}}
    \label{tab:rankme}
\end{table}

 Another simple way to evaluate SSL methods, called RankMe, was introduced by~\cite{garrido2022rankme}. The idea is to use the effective rank of representations, defined as the entropy of the singular value distribution of the embeddings. It can be computed as:
 \begin{equation}
    \text{RankMe}(\mZ) = \exp\left(-\sum_{k=1}^{\min(N,K)} p_k \log p_k\right),\; p_k = \frac{\sigma_k(\mZ)}{\|\sigma(\mZ)\|_1}+\epsilon
\end{equation}
 It is shown to be a necessary condition for good performance, though you can achieve full rank representations with degenerate results (e.g. a random matrix with entries sampled i.i.d. from a Gaussian distribution). While this cannot be used to evaluate different methods, it works well for hyperparameter selection, as shown in \Cref{tab:rankme}.

\begin{tikzpicture}
    \Quote{To debug without labels, looking at the rank of the representations is a good start e.g. with RankMe}{Quentin}
\end{tikzpicture}

\subsubsection{Going beyond classification}

While classification is a commonly used performance metric for evaluating self-supervised learning models, it is important to consider other types of vision tasks as well. Tasks such as object detection and semantic segmentation have gained popularity as they require models to learn more complex representations of visual information. Recent works~\cite{caron2021emerging, zhou2021ibot, bardes2022vicregl} have demonstrated the effectiveness of self-supervised learning for these tasks. However, a limitation is that there is currently no standardized protocol for evaluating self-supervised models on these tasks. Various evaluation methods exist, such as finetuning the encoder on a downstream task or using the encoder as a feature extractor. Further research is needed to establish a standardized evaluation protocol for these tasks in the context of self-supervised learning.

\subsubsection{Visual Evaluation}

\begin{figure}
\begin{center}
    \includegraphics[width=0.9\linewidth]{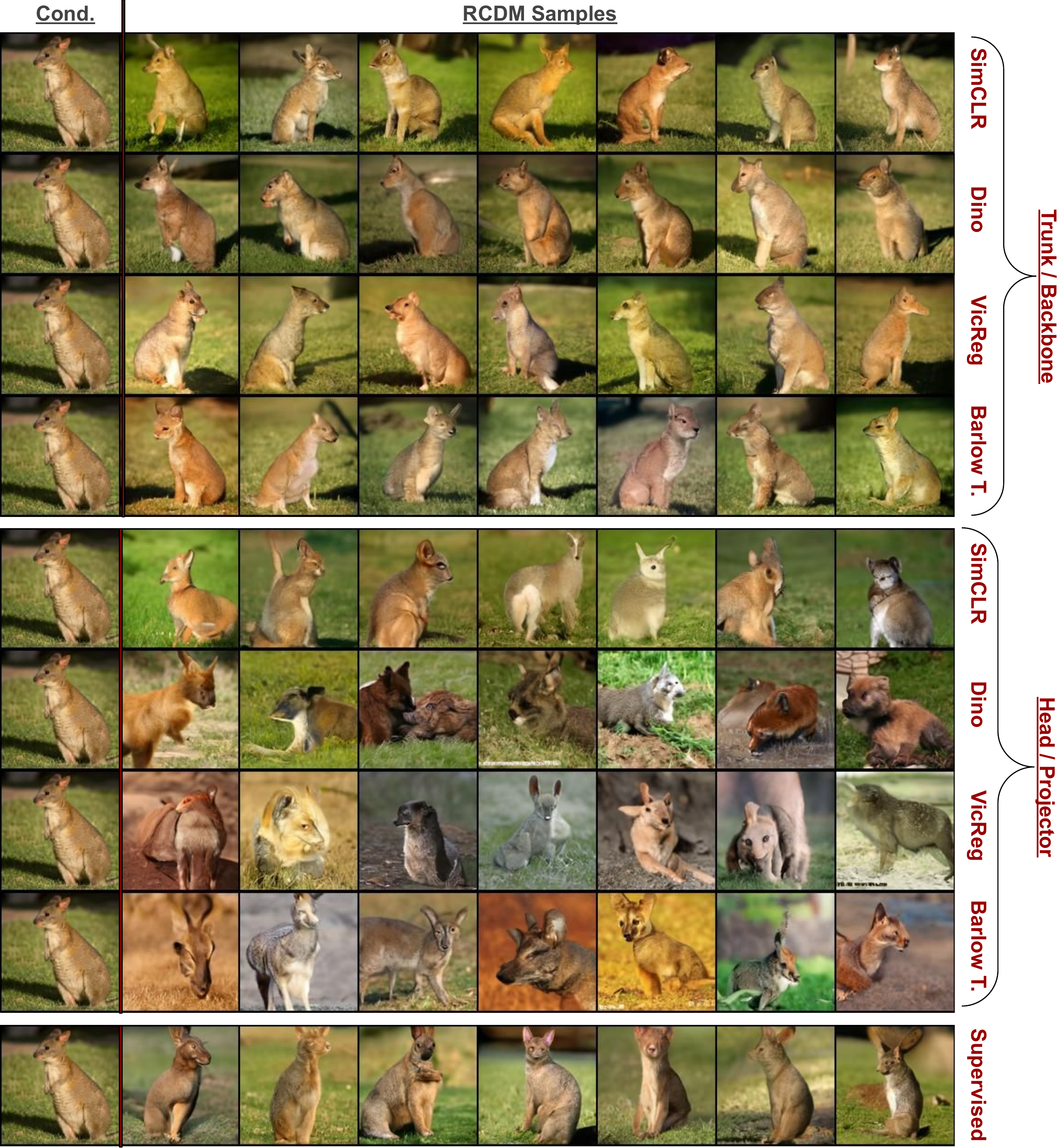}
    \caption{\small Figure from \citet{bordes2022high}. RCDM visualization of {\bf what is encoded inside various representations?} First to fourth rows show our samples conditioned on the usual resnet50 backbone representation (size 2048) while fifth to eigth rows show samples conditionned on the projector/head representation of various ssl models. (Note that a separate our generative model was trained specifically for each representation). \emph{Common/stable aspects} among a set of generated images reveal \emph{what is encoded} in the conditioning representation. \emph{Aspects that vary} show \emph{what is not encoded} in the representation. We clearly see that the projector representation only keeps global information and not its context, contrary to the backbone representation. This indicates that invariances in SSL models are mostly achieved in the projector representation, not the backbone. Furthermore, it also confirms the linear classification results of Table a) which show that backbone representation are better for classifications since they contain more information about an input than the ones at the projector level.}
    \label{fig:rcdm_proj_vs_backbone}
\end{center}
\vspace{-0.4cm}
\end{figure}

Another way to evaluate what information is contained or not in a representation is to use a decoder over the representation that is able to map back this information to pixel space. Some methods like \citep{he2022masked} are built with a specific decoder which make such visual analysis easy, however most SSL methods aren't shipped with a decoder. To alleviate this issue and to allow researchers to visualize what can be learned by any type of SSL method, \citet{bordes2022high} suggest training a conditional generative diffusion model using a SSL representation as conditioning. By analyzing which information remains constant across different generated samples using a specific conditioning and what information does not remain constant (because of the stochasticity in the generative model), one can get some hints about what information is contained in the representation. If a representation encodes every information about each pixel, the conditional generative model would exploit every bit of this information to perform a perfect reconstruction which will lead to no variance across different samples. If the representation encodes only the class information, the conditional generative model will only be able to use that to reconstruct the image belonging to this class, which means that when generating different samples, the object class will remain constant but the background/context/color would change across samples. In \Cref{fig:rcdm_proj_vs_backbone}, we show how RCDM was used by \citet{bordes2022high} to compare the representations learned at the projector level versus the representations learned at the backbone level. In this Figure, we observe that the representations at the projector level are much more invariant since the color/background information does not remain constant across different samples while this is not the case at the backbone level.

\begin{tikzpicture}
    \Quote{You often only need to train for a few epochs to test collapse (no more than 5 epochs)}{Adrien}
\end{tikzpicture}

\subsection{Speeding up Training}

\subsubsection{Distributed Training}

Training self-supervised models often requires large batch sizes~\citep{chen2020simple, He2020MomentumCF}, or can be considerably speed up by increasing the batch size, which is ultimately limited by the memory capacity of the device the model is trained on. Distributed training divides batches across several devices that run in parallel, which increases the overall size of the batch. 
This is mainly done with DDP: Distributed Data Parallel or FSDP: Fully Sharded Data Parallel, available in libraries like FairScale~\citep{FairScale2021} or Apex~\citep{apex}.
However some self-supervised methods rely on the statistics of the current batch for the computation of their loss value~\citep{chen2020simple, zbontar2021barlow, bardes2021vicreg}, which has to be taken into account when distributing the training across multiple devices. In this section, we present the elements that need to be taken into account in order to correctly distribute the training of common self-supervised learning methods. We call effective batch size, the size of the full batch distributed on the devices, and per device batch size, the size of each sub-batch on a single device.

{\bf Synchronized batch normalization.} Batch normalization is one a the most common technique for stabilizing neural network training, as well as improving the performance of the network. It is present in most convolutional backbones used in self-supervised learning, in particular in ResNet. Batch norm uses the statistics from the current batch, which need to be aggregated for distributed training. This can be done easily in PyTorch by wrapping your distributed model the following way:
\texttt{model = torch.nn.SyncBatchNorm.convert\_sync\_batchnorm(model)}
This will replace all the BatchNorm modules in the network by a custom BatchNorm class that aggregates the statistics automatically.

{\bf Aggregate batches for exact loss computation.} Batch norm is not the only operation that operate on batches, multiple self-supervised loss functions do as well, such as SimCLR~\citep{chen2020simple} that uses the examples in the current batch as negative example for its contrastive loss, or VICReg~\citep{bardes2021vicreg} that computes the covariance matrix of its embeddings. In these cases the batches from each device need to be aggregated into the full batch manually. This can be done using the all\_gather operation from PyTorch, however this operation does not allow back-propagation through it. We therefore implement a custom gather operation that does, the code is provided below:

\begin{algorithm}
  \caption{}
  \label{alg:method}
    \definecolor{codeblue}{rgb}{0.25,0.5,0.5}
    \definecolor{codekw}{rgb}{0.85, 0.18, 0.50}
    \newcommand{\algofontsize}{8.5pt}
    \lstset{
      backgroundcolor=\color{white},
      basicstyle=\fontsize{\algofontsize}{\algofontsize}\ttfamily\selectfont,
      columns=fullflexible,
      breaklines=true,
      captionpos=b,
      commentstyle=\fontsize{\algofontsize}{\algofontsize}\color{green!50!black},
      keywordstyle=\fontsize{\algofontsize}{\algofontsize}\bfseries\color{blue!90!black},
    }
\begin{lstlisting}[language=python]
class GatherLayer(torch.autograd.Function):
    """
    Gather tensors from all process and support backward propagation
    for the gradients across processes.
    """

    @staticmethod
    def forward(ctx, x):
        output = [torch.zeros_like(x) for _ in range(dist.get_world_size())]
        dist.all_gather(output, x)
        return tuple(output)

    @staticmethod
    def backward(ctx, *grads):
        all_gradients = torch.stack(grads)
        dist.all_reduce(all_gradients)
        return all_gradients[dist.get_rank()]
\end{lstlisting}
\end{algorithm}

We use an \texttt{all\_reduce} operation on the gradient, which sums them, because DDP will divide them later by the number of devices. One can use the operation by simply calling: \texttt{FullGatherLayer.apply(x)} on the input \texttt{x}. Practically, for the methods above, this needs to be done on the embeddings just before the computation of the loss.

{\bf Additional tricks.} We advise to always use the effective batch size as argument to the training script, as well as for comparing runs. The \texttt{DataLoader} class takes the per device batch size as, argument, which can be obtained by dividing the effective batch size by the number of devices which is \texttt{world\_size} in PyTorch. We also advise to use an adaptive learning rate scaled with the effective batch\_size, for example using \texttt{effective\_lr = base\_lr * effective\_batch\_size / 256} where \texttt{is the base\_lr} is the argument of the training script. This reduce the learning rate search range when changing batch\_size. When using small batch size, it is recommended by \citet{chen2020simple} to use \texttt{effective\_lr = base\_lr * $\sqrt(\text{effective\_batch\_size})$ / 256}.

\label{sec:ffcv}
\subsubsection{Even Faster Training with FFCV and Other Speedups}
\begin{figure}
    \centering
    \includegraphics[width=\linewidth]{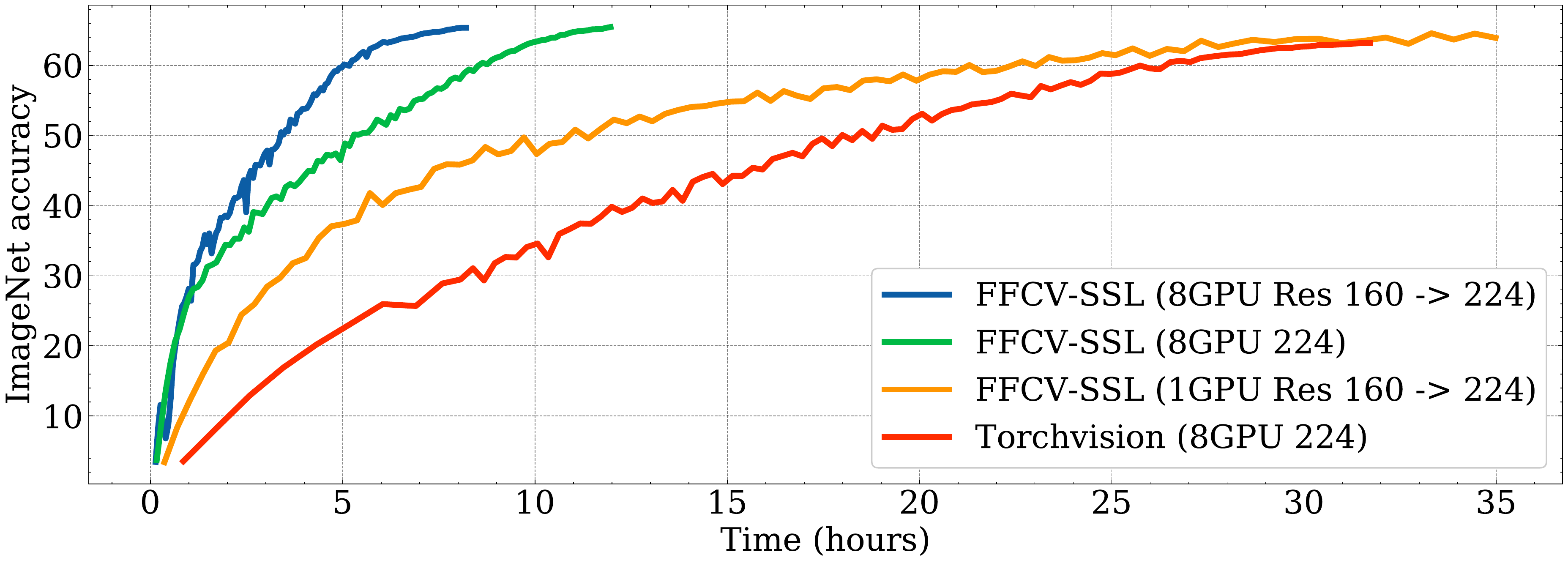}
    \caption{Figure from \citet{demo}. ImageNet validation accuracy (y-axis) during training of SimCLR with respect to the training time (x-axis). FFCV-SSL is a library that is specifically optimized for Self-Supervised Learning, and that extends the original FFCV library \citep{leclerc2022ffcv}. FFCV-SSL allows a 3x time speed up with respect to torchvision and enables the training of SSL model in less than 2 days on a single gpu. }
    \label{fig:ffcv_vs_torchvision}
\end{figure}

Since most join-embedding SSL methods requires different set of handcrafted data augmentation, data processing can become a real bottleneck when training SSL models. Some approaches\footnote{\url{https://github.com/vturrisi/solo-learn}} have used DALI as an alternative data loader to pytorch vision while some other have relied on FFCV-SSL\footnote{\url{https://github.com/facebookresearch/FFCV-SSL}} which is based on the FFCV library~\citep{leclerc2022ffcv}. FFCV-SSL~\citep{demo} shows that one can train SimCLR on ImageNet in less than 2 days on a single GPU or in just a few hours using 8 GPUs (Figure \ref{fig:ffcv_vs_torchvision}).

\subsubsection{Speeding Up Training of Vision Transformers}

Training ViT can be made more efficient for two reasons. 
First, it is made easy for ViTs not to process all patches.
This is especially helpful when using masked prediction pre-training objectives such as MAE~\citep{he2022masked} or Masked Siamese Networks~\citep{assran2022masked}.
For instance, with ViT and such objectives, Data2vec 2.0~\citep{baevski2022efficient} achieves $84\%$ top-1 accuracy after pre-training for only $3$ hours on $32$ GPUs.

The second reason is linked to the architecture.
Since transformers~\citep{vaswani2017attention} are employed in almost all domains of computer science, many works aim to reduce the compute and memory requirements of the attention mechanism.
One approach is with low-rank and/or sparse approximation mechanisms~\citep{kitaev2020reformer, choromanski2020rethinking, wang2020linformer, chen2021scatterbrain, zaheer2020big}. For instance, \cite{li2022efficient} use sparse self-attention to improve efficiency in the context of SSL vision models.
Another approach, is to resort to IO-aware optimizations~\citep{ivanov2021data}, the most known one perhaps being FlashAttention~\citep{dao2022flashattention}.

These speed-ups are available in open-source libraries: Fairseq~\citep{ott2019fairseq}, FairScale~\citep{FairScale2021}, XFormers~\citep{xFormers2022}, Apex~\citep{apex}, etc.

Another simple way to speed up the training of vision transformers is to use Pytorch bfloat16 which allow faster training while keeping the same precision range as float32 (this is useful to avoid the usual numerical instability issues one can encounter when training vision transformers in float16).

\section{Extending Self-Supervised Learning Beyond Images and Classification}

\subsection{Strategies for Other Data Domains}

Pre-training large models with self-supervision objectives is popular not only for vision systems, but also for audio, text, and tabular data as well.  The performance of existing SSL methods varies across these domains -- yielding state-of-the-art language models but limited success on tabular data -- which may either reflect better suitability of self-supervision or alternatively the wildly differing amount of attention which has been paid to the various domains in the SSL literature.

Applying SSL techniques to any of these data domains requires care as unique challenges arise in each domain which necessitate special considerations.
For example, SSL for vision often revolves around data augmentations that may not naturally apply to speech signals. The `positive pairs' available for contrastive learning varies from slightly different views of the same image to totally different segments of an audio recording. 
Nonetheless, both contrastive and generative objectives can be applied to these other data domains.  One generically useful technique across data types is masking.  Whether predicting missing words in a sentence, pixels in an image, or entries of a row in a table, masking is an effective component of SSL approaches across domains.

This section is not intended as a thorough survey of self-supervision for other data modalities, as each of those fields is vast.  Domain-specific surveys can be found in \citet{liu2022audio} (audio),  \citet{schiappa2022video} (video), \citet{min2021recent} (text), and \citet{rubachev2022revisiting} (tabular data). 
Rather, this section provides a discussion of the interesting similarities and differences in how SSL is applied to audio, text, and tabular data.

\textbf{Audio data.} Audio signals, both raw audio and mel spectrograms, have a lot in common with images. 
As inputs to a neural network, there are strong similarities. For example convolutions can be useful \citep{oord2016wavenet, schneider2019wav2vec, baevski2021unsupervised}. 
But as data for SSL, major differences arise. 
For example, horizontally flipping an image does not usually change the semantic meaning of an image (and is a wildly popular data augmentation), but for speech recordings this would completely distort the data. 
Similarly, while masking images is often done with random pixels, the two dimensions of a spectrogram represent time and frequency and masking with horizontal and/or vertical bands is more effective \citep{wang2020unsupervised}.
Additionally, the existence of tones other than speech (background noise, room tone) presents a unique challenge when looking for positive pairs for contrastive learning, which is to prevent the learned representations from over fitting to the noise withing a given clip \citep{oord2018representation, wang2020unsupervised}.
In fact, the high frequency noisy artifacts, which are generally unrelated to the semantic meaning, mean reconstruction in input space is more complicated than other domains (e.g. text). 
Multi-modal models, on the other hand, can consider a soundbite and its text \citep{sermanet2018time, chung2018unsupervised} or some frames of a video and the corresponding sound clip \citep{zhao2018sound, alwassel2020self} as different views to be used as positive pairs for contrastive learning.

\textbf{Video data.}
Most of SSL images methods have a counter-part video SSL method.
For instance, \cite{feichtenhofer2021large} have generalized SimCLR, MoCo, SwAV and BYOL to space-time video data. Indeed, in all these methods it is possible to incorporate the notion of similarity between different temporal clips of the same video.
More recently, masked auto-encoding objectives for video have been built around the same idea as images, but by masking patches/ tubes of patches in the temporal axis as well~\citep{feichtenhofer2022masked, tong2022videomae, girdhar2022omnimae}.
Besides, it is common practice to use SSL vision pre-trained models for video downstream tasks like action recognition.
With ViT for instance, the patch embedding convolutional layer can be transferred from 2D to 3D by repeating the weights along the temporal axis~\citep{feichtenhofer2022masked}. Vision models can then transfer to video models by using them as initialization for fine-tuning on video tasks~\citep{fang2022eva}. 
The frame features can also be used directly, by appending a linear layer on top of the features~\citep{radford2019language}, or by using more complex heads~\citep{ni2022expanding, arnab2021vivit}. In this case, the visual system is frozen and the temporal information is learned after.

\textbf{Text data.} 
In contrast to audio data, text is a relatively clean input signal and representations that are useful for reconstruction do not over fit to a noisy part of the signal. 
In fact, the most popular large language models are all trained with reconstruction objective as opposed to contrastive objectives popular in other data domains \citep{radford2018improving, radford2019language, brown2020language, devlin2018bert}. The Word2Vec objective \citep{mikolov2013distributed} predicts a masked out portion of the training text has served as a foundational objective for self-supervised learning in natural language. While uncommon, language modeling can be done with contrastive learning for word or or character representations \citep{chen2022clower}.
One other difference between text and images is that the masked token prediction for text is done over an entire dictionary. This approach is not the dominant one for images but it has been tried at the pixel level \citep{chen2020generative}. 
While there are few augmentations for language data that do not change the semantic meaning, large scale systems generally use enough data and various types of masking to overcome this. 
Specifically, next token prediction \citep{radford2018improving, radford2019language, brown2020language} is akin to masking the last token in a string, while bidirectional encoders mask tokens anywhere in the string \citep{devlin2018bert} or fill larger spans of missing text \citep{raffel2020exploring,tay_unifying_2022}. This choice of unidirectional next-token prediction versus bidirectional approaches leads to meaningful differences in downstream text applications \citep{artetxe_role_2022}.
For contrastive learning, positive pairs often come from masking and/or cropping input sequences \citep{meng2021coco, giorgi2021declutr}. 
They can also be generated using dropout so that one input has two different latent representations \citep{gao2021simcse}.
Additionally, some methods for both contrastive and reconstructive pretraining corrupt the input with several other augmentations including document rotation, sentence permutation, and token deletion \citep{lewis2020bart, raffel2020exploring, wu2018unsupervised}. 

\textbf{Tabular data.} 
Unlike text, audio, and images, classical machine learning tools are still popular for processing tabular data.
However, while deep learning for tabular data is comparatively a small field, finding sensible data augmentation strategies is already a much studied topic. 
Several SSL methods for tabular data utilize masking in various ways and some techniques creatively employ other augmentations developed for images, like mixup \citep{zhang2018mixup}. As with images and audio, some algorithms aim to generate the missing or corrupted values while others employ contrastive learning. In combinatorial optimization such as Mixed Integer Programming (MIP), objective function is used as the guidance to generate positive solution pairs with comparable objectives and negative solution pairs whose objective values drastically differs despite tiny changes of a few variables~\citep{huang2023searching}. Similar approaches are also used in guided language generation~\citep{yang2022doc}. 

The masked reconstruction approaches account for a variety of masking tactics.
Furthermore, it is common with tabular data to predict mask vectors as a pretext task \citep{yoon2020vime, iida2021tabbie}. 
Since the predicting mask itself is part of the pretraining objective, the masked entries in the input must be filled, and typically this is done by sampling from the empirical distribution of that column or feature.

With the same augmentation, i.e. masking and sampling from the empirical marginal distribution, \citet{bahri2021scarf} propose pretraining with a contrastive loss. Specifically, they propose using the InfoNCE loss \citep{gutmann2010noise, ceylan2018conditional} to compare the representations of the clean and corrupted inputs.

Several other works outline ways to augment the data for a combination of generation and contrastive learning. 
For example, tabular data can be split into groups of columns so each sample (row) has several views available \citet{ucar2021subtab}. 
Borrowing from vision systems, a combination of CutMix \citep{yun2019cutmix} in input space and mixup \citep{zhang2018mixup} in embedding space is also an effective augmentation for tabular data \citep{somepalli2021saint}. 
These methods generate augmented views that are used along with the clean input for contrastive learning. 
However, contrastive pretraining for both the SAINT model \citep{somepalli2021saint} and SubTab \citep{ucar2021subtab} seems to work best when this is paired with a reconstructive loss term.

In their work focusing on comparing the SSL methods for tabular data, \citet{rubachev2022revisiting} find that pretraining objectives generally do help boost the performance of tabular models. But more specifically, they find that pretraining objectives that use the labels are best, implying that SSL for tabular data has yet to be the state of the art in its domain \citep{rubachev2022revisiting}. Similarly, \citet{levin2023transfer} show that unlike in computer vision, existing SSL pre-training routines yield less transferable features than supervised pre-training.

\textbf{Reinforcement learning.}
SSL has been used to improve reinforcement learning (RL) on visual inputs. This setting is similar to video, except apart from the sequence of images, we also have access to the sequence of actions. The most common approach to apply SSL here is to use contrastive learning to train a model to match current state representation and the next time step's representation, or to match representations of the same state but with different augmentations applied. One of the earliest examples is CURL \citep{Srinivas_Laskin_Abbeel_2020}. Recently, SSL has been used to improve sample efficiency on a challenging Atari100k benchmark \citep{simple}. Recent works have modified BYOL \citep{grill2020bootstrap} or Barlow Twins \citep{zbontar2021barlow} by feeding images of consecutive timesteps' observations to the two branches of the siamese network: SGI \citep{sgi} and Barlow Balance \citep{Zhang_GX-Chen_Sobal_LeCun_Carion_2022} did this for offline pretraining, while SPR \citep{spr} uses it as an additional objective in the online setting. The best-performing method doing this is EfficientZero \citep{Ye_Liu_Kurutach_Abbeel_Gao_2021}, which modifies MuZero \citep{muzero} by, among other modifications, adding the SimSiam \citep{Chen_He_2020} objective to train the encoder and the forward model, and sets the new state of the art on Atari100k. \citet{Parisi_Rajeswaran_Purushwalkam_Gupta_2022} propose PVR, a method based on MoCo \citep{he2020momentum} that improves sample efficiency on control tasks. \citet{Eysenbach_Zhang_Salakhutdinov_Levine_2022} show that contrastive learning in RL setting is directly linked to goal-conditioned RL, and demonstrate that a method based on InfoNCE \citep{oord2018representation} achieves great performance on robotic arm control tasks.

SSL has been shown to yield good representations for behavior cloning. \citet{Pari_Shafiullah_Arunachalam_Pinto_2022} show that imagenet-pretrained model finetuned with BYOL \citep{grill2020bootstrap} can be very effectively used for visual imitation on robotic open, push and stack tasks, while \citet{Arunachalam_Guzey_Chintala_Pinto_2022} use a similar method and successfully learn from a small manipulation dataset collected using VR. \citet{tdex} present a method that uses BYOL to extract information from tactile sensors on robotic arms and improve dexterous manipulation. \cite{cbet} shows that BYOL representations of visual inputs are also useful when modeling goal-conditioned trajectories with transformer architecture.

There are a few additional challenges when applying SSL to RL. First, if the data is recorded on-line, individual observations are highly correlated with each other and are not IID (independent and identically distributed), so sampling from replay buffer should be done carefully. One failure mode of SSL objectives when applied to RL agents' data is the proclivity to latch on `slow features' \citep{slowfeats}. The contrastive objective may learn for example to only look
at the cloud patterns in the sky to tell apart frames in a self-driving dataset, so one must be careful to design augmentations in a way to 
remove useless static features in the image, or to sample data accordingly.

SSL has been used not only to improve sample efficiency, but also to improve exploration. \citet{byolexplore} propose BYOL-Explore which uses BYOL \citep{grill2020bootstrap} to learn the encoder and the forward model, and use the forward model disagreement as the exploration objective. The follow-up work by \citet{Jarrett_Tallec_Altché_Mesnard_Munos_Valko_2022} address the problem of BYOL-Explore latching on a noisy TV. \citet{Yarats_Fergus_Lazaric_Pinto_2021} proposed using a clustering method akin to SwAV \citep{caron2020unsupervised} to do unsupervised exploration, i.e. exploration with only intrinsic rewards.

A few works have explored using vast natural videos data available to pre-train representations for RL agents.
\citet{Xiao_Radosavovic_Darrell_Malik_2022} introduce MVP, which uses masked-autoencoder to pre-train the transformer encoder for robotic control, while \citet{Ma_Sodhani_Jayaraman_Bastani_Kumar_Zhang_2022} propose VIP, a method to learn universal features for RL using ResNet-50 backbone and the objective based the time between frames in the observations as the supervision signal. Another method for training foundation models for RL, R3M \citep{Nair_Rajeswaran_Kumar_Finn_Gupta_2022}, combines time-contrastive and video-language alignment objectives. VIP and R3M are trained on the large Ego4D dataset \citep{ego4d}, while MVP combines Imagenet, Ego4D, and additional hand manipulation data. \citet{Majumdar_Yadav_Arnaud_Ma_Chen_Silwal_Jain_Berges_Abbeel_Malik} propose VC-1, a method based on masked auto-encoding. The authors test the proposed method and other foundation models on
the new test suite called CortexBench. The benchmark includes control, object manipulation, and navigation tasks, with different methods excelling at different parts of the benchmark.

There are also unsupervised methods for learning representations that are specific to RL and are not commonly used for images: e.g. Laplacian eigenmaps \citep{Machado_Bellemare_Bowling_2017}, forward-backward representations \citep{Touati_Rapin_Ollivier}. \citet{Zhang_McAllister_Calandra_Gal_Levine_2021} propose to learn representations by making representations the same for states that lead to the same rewards, and different otherwise.

\subsection{Incorporating Multiple Modalities into SSL Training}
\label{subsec:multimodal}

Self-supervised learning need not be based on a single modality. Especially multi-modal vision-language have recently demonstrated this to great effect. Contrastive Language–Image Pre-training (CLIP) \citep{radford_learning_2021}, and ALIGN \citep{jia_scaling_2021} are self-supervised learning approaches that use image-caption pairs to learn a joint embedding space for images and captions. The objective here is contrastive, given an image and its caption are fed through separate encoder models that encode each modality into a fixed-length embedding vector. The embeddings of the training data image-caption pair are aligned, whereas other combinations in a batch are repelled. 

This approach is especially interesting in comparison to contrastive SSL based on pure vision, as discussed in \Cref{sec:contrastive}. The use of a second modality, here text, anchors the entire SSL training. It is no longer necessary to generate multiple augmented views to form a notion of robust representation as the joint approach learns semantically meaningful representations simply by observing similar captions re-occurring with similar images. 

As a result, image encoders arising from such a joint pre-training are especially robust to visual changes that leave semantic meaning unchanged, such as sketches of objects as evaluated in ImageNet-Sketch \citep{wang_learning_2019,radford_learning_2021}, and are strong on out-of-domain generalization tasks. Yet, this is not always a desired representation, as visualizations in \citet{ghiasi_what_2022} show that these models also group features that are visually dissimilar, but semantically, or literally, alike. 
This can be mitigated, and overall performance, e.g. in linear probing, can even be improved by combining both image-text and image-image SSL as done in \citet{mu_slip_2022}, who combine CLIP and SimCLR \citep{radford_learning_2021,chen2020simple}.

Recent work has pushed these vision-language systems to larger scales  \citep{ding_cogview_2021,yuan_florence_2021,singh_flava_2022,wang_simvlm_2022,fang_eva_2022}, based on freely available image-caption pairs collected from the internet, such as in \citep{schuhmann_laion-5b_2022}. These modern SSL models are capable of representing both vision and text, and can be used in a number of applications that are multimodal, from visual-question answering to multimodal generation \citep{alayrac_flamingo_2022,li_blip_2022,nichol_glide_2022,rao_denseclip_2022}.

The future of vision-language pre-training, as an alternative to robust visual representations learned on vision alone, remains to be further explored. While its advantages in vision-language downstream applications are evident \citep{shen_how_2022,dou_empirical_2022}, shared embedding spaces can also be constructed by training solely the vision encoder first, fixing it, and then training a matching language encoder, as described in \citep{zhai_lit_2022}. 
Ultimately, vision-language models are only the first step to self-supervised learning from multiple modalities at scale. Prototypes, such as \citet{reed_generalist_2022}, train self-supervised on arbitrary input streams, ranging from vision and text to tables and agent actions, and so learn re-usable representations that are helpful for general tasks.

\subsection{Building Feature Extractors with Localization for Dense Prediction Tasks}

Aside from semantic understanding, popular computer vision tasks from object detection to segmentation to depth estimation require models which extract localized features, in other words ones which contain information indicating the locations of objects within the input image.  Self-supervised learning may be particularly valuable for these dense prediction tasks since collecting segmentation masks or bounding box annotations for training images is significantly more expensive than classification labels.  However, learning frameworks which are carefully tuned on image classification benchmarks may lack traits which are valuable for such dense prediction tasks.  Several works, which we note perform their experiments in different settings and on different architectures and learning algorithms, express seemingly contradictory findings, namely that existing self-supervised learning strategies are or are not effective for downstream dense prediction tasks \citep{goyal2019scaling, purushwalkam2020demystifying, zhao2021distilling, ericsson2021well, shwartzpre}.  We now delve further into this discussion.

\textbf{Limitations of self-supervised learners for localization. }  SSL approaches which rely on augmented views or jigsaw transformations, such as MoCo \citep{He2020MomentumCF} and PIRL \citep{misra2020self}, learn occlusion invariance since they are trained with random crops on ImageNet where foreground objects are often large so that different crops contain different parts of the same object \citep{purushwalkam2020demystifying}.  On the other hand, they lack viewpoint invariance and category-instance invariance.  Further, \citet{zhao2021distilling} argue that self-supervised learners also lack localization information because the models are able to use all parts of the image, both foreground and background, to make their predictions.  The above works conduct experiments principally on convolutional architectures. 
It is worth noting that \citet{ericsson2021well} suggest that the best among the popular SSL algorithms they test on are CNNs, which can still achieve competitive performance with their supervised learning counterparts in some detection and segmentation settings.  
Interestingly, older pretext tasks such as \texttt{jigsaw} or \texttt{colorization}, which predate the recent SSL craze sparked by MoCo and SimCLR, can also achieve competitive performance compared to supervised learning backbones when the pretext task is made ``hard’’ enough \citep{goyal2019scaling}.

\textbf{CNNs or ViTs? } Recent works suggest that vision transformers (ViTs) contain superior localization information in their learned representations compared to convolutional architectures \citep{caron2021emerging}.  Whereas CNNs require specially designed segmentation pipelines to extract localization information from their features, this information arises naturally in the patchwise features of ViTs.  Existing SSL methods designed specifically for transformers confirm that the trained models are effective for downstream detection and segmentation tasks, especially when fine-tuned \citep{li2021mst, he2022masked}.  However, it should be noted that these SSL algorithms explicitly demand localization in their objective functions, for example via masked autoencoding where patch features should contain information regarding the contents of the corresponding section of the image \citep{he2022masked}.  More recently, masked autoencoding pre-training strategies have been adapted for convolutional architectures to great effect, where they achieve competitive performance on downstream object detection and instance segmentation \citep{woo2023convnext}.  Moreover, we will see below that a variety of pre-training strategies designed specifically for localization can be effective on transformers and convolutional networks alike.

\textbf{So how do we learn localized features without annotations? } In order to tailor representations for downstream dense prediction tasks, numerous works propose modifying SSL routines specifically to enhance the localization in their features.  Since these SSL pre-training algorithms do not use segmentation or detection annotations, they instead rely on carefully chosen unsupervised object priors.  

One style of object prior enforces relationships between features extracted from locations within a single image, just as self-supervised learning procedures often enforce relationships between distinct images.  One such prior uses the fact that adjacent ViT patches often contain the same objects.  Unlike popular contrastive objectives which encourage augmented views of an image to produce similar features, SelfPatch encourages adjacent patches within a single image to produce similar features \citep{yun2022patch}.  A related method, DenseCL \citep{wang2021dense}, matches the most similar pixel-wise features extracted from augmented samples to automatically handle the case in which augmentations move objects around in an image, and we only want to match features corresponding to the same object. More recently, VICRegL \citep{bardes2022vicregl} applies a similar principle by combining geometric and learned matching, with a non-contrastive criterion. Just as clustering-based methods cluster related images, Leopart \citep{ziegler2022self} fine-tunes a pre-trained model to cluster patch-level features.

In addition to modifying the training loss to improve localization, we can also augment the data with this objective in mind by placing an object in multiple settings so that resulting models extract the same features from an object irrespective of its location.  Instance Localization \citep{yang2021instance} leverages RoIAlign \citep{he2017mask}, an algorithm designed for object detectors which extracts features corresponding to a specific image patch.  To this end, Instance Localization pastes a randomly chosen patch cut from the foreground of one image onto two other images and extracts features corresponding to only the pasted foreground patch, using a contrastive loss to ensure that the foreground patch generates similar features regardless of the background present and regardless of its location within an image.  A competing approach estimates the location of an object within the training image using saliency maps and then cuts and pastes these objects onto background and optimizes a similar objective \citep{zhao2021distilling}.  Instead of using augmentations to move objects around, \citet{purushwalkam2020demystifying} notes that nearby video frames contain the same object but in different positions or from different viewpoints so that contrastive learning on video data can serve much the same purpose.

Recently, UP-DETR~\citep{dai2021up} and DETReg~\citep{bar2022detreg} proposed an end-to-end SSL pretraining of the DETR family detectors. UP-DETR proposes to detect the bounding boxes of randomly selected patch regions in images conditioned on their pixel values while predicting their corresponding SwAV~\citep{caron2020unsupervised} embedding. In DETReg, detection targets are obtained using the Selective Search algorithm, which does not require human annotations. Similarly, the detector predicts an associated SwAV~\citep{caron2018deep} embedding for each target bounding box.

\textbf{Vision-language models for dense prediction tasks. } In \Cref{subsec:multimodal}, we saw that vision-language models extract semantically meaningful features. These features are also leveraged by recent works for open-vocabulary object detection \citep{kamath2021mdetr, gu2021open, zareian2021open, minderer2022simple}.  These works leverage vision and language backbones pre-trained as previously discussed on captioned image databases and fine-tune on object detection data.  Crucially, pre-trained language models, paired with image feature extractors, allow open-vocabulary object detectors to detect new objects never seen during their fine-tuning stage simply by querying the language model with an appropriate prompt.

\section{Conclusion}

\textit{Self-supervised learning} (SSL) established a new paradigm for advancing machine intelligence.
Despite many successes, SSL remains 
a daunting field with 
a dizzying array of methods each with intricate implementations. Due to the fast moving research and the breadth of SSL methods, it remains a challenge to navigate the field. This becomes an issue for researchers and practitioners who joined the field only recently, in turn creating a high barrier to entry for SSL research and deployment.
We hope our cookbook will help lower these barriers by enabling the curious researcher of any background to navigate
the terrain of methods, understand the role of the various knobs, and gain the
know-how required to be successful with SSL.

\newpage
\bibliography{references}
\bibliographystyle{abbrvnat}

\end{document}